\documentclass[sn-mathphys-num]{sn-jnl}

\usepackage{graphicx}
\usepackage{subcaption}
\usepackage{multirow}%
\usepackage{amsmath,amssymb,amsfonts}%
\usepackage{amsthm}%
\usepackage{mathrsfs}%
\usepackage[title]{appendix}%
\usepackage{xcolor}%
\usepackage{textcomp}%
\usepackage{manyfoot}%
\usepackage{booktabs}%
\usepackage{algorithm}%
\usepackage{algorithmicx}%
\usepackage{algpseudocode}%
\usepackage{listings}%
\usepackage{natbib}%
\usepackage{booktabs}
\usepackage{tabularx}

\begin{document}

\title[Augmented Functional Random Forests]{\textbf{Augmented Functional Random Forests: Classifier Construction and Unbiased Functional Principal Components Importance through \textit{Ad-Hoc} Conditional Permutations}}

\author*[1]{\fnm{Fabrizio} \sur{Maturo}}\email{fabrizio.maturo@unimercatorum.it}

\author[2]{\fnm{Annamaria} \sur{Porreca}}\email{annamaria.porreca@unimercatorum.it}
\equalcont{These authors contributed equally to this work.}

\affil*[1]{\orgdiv{Department of Economics, Statistics and Business}, \orgname{Faculty of Technological and Innovation Sciences}, \orgaddress{\street{Piazza Mattei, 10}, \city{Rome}, \postcode{66100}, \country{Italy}}}

\affil[2]{\orgdiv{Department of Economics, Statistics and Business}, \orgname{Faculty of Economics and Law}, \orgaddress{\street{Piazza Mattei, 10}, \city{Rome}, \postcode{66100}, \country{Italy}}}

\abstract{This paper introduces a novel supervised classification strategy that integrates functional data analysis (FDA) with tree-based methods, addressing the challenges of high-dimensional data and enhancing the classification performance of existing functional classifiers. Specifically, we propose augmented versions of functional classification trees and functional random forests, incorporating a new tool for assessing the importance of functional principal components. This tool provides an ad-hoc method for determining unbiased permutation feature importance in functional data, particularly when dealing with correlated features derived from successive derivatives. Our study demonstrates that these additional features can significantly enhance the predictive power of functional classifiers. Experimental evaluations on both real-world and simulated datasets showcase the effectiveness of the proposed methodology, yielding promising results compared to existing methods.}




\maketitle


\section{Introduction}
\label{sec1:intro}

Managing vast amounts of data from sources like phones, sensors, and calculators has become crucial in today's data-driven world. Technological advancements have led to the development of tools for handling this data, used in areas such as environmental control, healthcare, and more. As a result, dimensionality reduction and classification techniques are increasingly important in fields like medicine, multimedia processing, and robotics. However, learning from high-dimensional data is challenging due to irregular time points, computational complexity, balancing bias and variance, and improving model interpretability and performance metrics. The curse of dimensionality, which complicates data analysis, is a well-known challenge. To address these issues, functional data analysis (FDA) has become a widely adopted approach \citep{Ramsay2002, Ferraty2003, Ramsay2005, Febrero2012}.

Functional Data Analysis (FDA) is a statistical domain focused on applying statistical methods to data represented as functions rather than traditional real numbers or vectors. By treating functions as unified entities, FDA introduces a new paradigm in statistical modeling and prediction. The benefits of FDA, such as utilizing derivatives for better insights, adopting non-parametric approaches, reducing dimensionality, and exploiting pattern sources, are well-documented in recent literature \citep{Ramsay2002, Ferraty2003, Ramsay2005}. The methodological advancements in FDA aim to bridge traditional statistics with functional approaches, and FDA tools are increasingly used in high-dimensional time-series analysis across various fields \citep[see e.g.][]{Ramsay2005, Ferraty2006, Aguilera2013, maturo2019fuzzy}. A key method in FDA is the functional principal component decomposition (FPCD), which represents functions through a linear combination of functional principal components (FPCs) to maximize variance, facilitating dimensionality reduction while retaining essential information \citep{Ramsay2002, Ferraty2003, Ramsay2005, Cuevas2014}. This approach has expanded conventional statistical methods to functional data, particularly in supervised classification, the focus of this paper.

Supervised classification in FDA focuses on developing classification rules from observed curves to predict the classes of new curves with high accuracy. Several methods have been explored, including Logistic Classifier, k-nearest Neighbour Classifier, and Kernel Classifier \citep[see, e.g.][]{Cuevas_2007, Preda_2007, AguileraMorillo2012, Escabias2014, Cuevas2014, Gregorutti_2015}. Recent research has increasingly combined FDA with tree-based techniques for classification tasks. For example, \citep{Yu_1999} introduced spline trees for functional data, focusing on time-of-day patterns for international call customers. \citep{Nerini_2007} developed a regression tree within the FDA framework for probability density functions. \citep{Gregorutti_2015} explored variable importance in FDA combined with tree methods, while \citep{Moller2016} proposed random forests for functional covariates using mean values over time windows. \citep{El_Haouij_2018} extended random forests with wavelet basis for classifying driver's stress levels. Further, \citep{Rahman_2019} developed a classifier for dose-response predictions with curve outcomes. \citep{maturo2023supervised} and \citep{maturo2024combining} explored using functional principal components for classification trees and combining clustering with tree-based classification, respectively. Lastly, \citep{Maturo2022SIM} proposed an innovative evaluation of leaf quality in functional classification trees applied to biomedical signals with binary outcomes.

Given that the starting point is undoubtedly to take advantage of the blended use of FDA \citep{Ramsay2002, Ferraty2003} and statistical learning techniques \citep{Hastie2009}, it is evident that, in this framework, much must still be discovered and tested for managing vast quantities of data and trying to interpret the results from a statistical perspective, possibly also from a causal and inferential viewpoint. Hence, investigations in this field are currently at the forefront, dynamic, and full of potential. In the coming years, we will observe many articles that enhance functional classifiers' precision, interpretability, and explainability.

Based on what has been introduced above and the many possible research scenarios this topic offers, this paper presents novel supervised classification strategies called Augmented Functional Classification Tree (AFCT) and Augmented Functional Random Forests (AFRF). Starting from the basic idea proposed in \citep{Maturo2022SIM}, this study aims to tackle the challenge of learning from high-dimensional data and enhance classification performance and models' explainability. This article proposes to exploit additional features of the original functional data to improve the information fed to the functional classification tree ensembles.
To steal a typical machine learning terminology adopted for image recognition, in this study, we talk about ``\textit{augmented functional data}'' to suggest the joint use of sequential derivatives to extract new features for supervised classification. In other words, this approach can be seen as observing functions from different perspectives to capture additional features that can help improve classification performance. 
Additionally, we introduce a novel method called Conditional Permutation Importance for Augmented Functional Principal Components (CPIAFPC). This method provides an unbiased assessment of feature importance by accounting for the inherent correlations among features derived from successive derivatives. CPIAFPC ensures that the contribution of each feature to the classification model is accurately represented, further enhancing the interpretability and reliability of the proposed classification strategies.

Extensive experimental evaluations are conducted on real-world and simulated datasets to evaluate the effectiveness of the proposed methodology. The comparisons to existing methods show exciting results in terms of classification performance. The research contributes to the field of FDA by demonstrating its potential for handling high-dimensional data in supervised classification tasks and offering valuable insights into the underlying functional characteristics of the data.

The paper proceeds with Section 2, introducing the fundamental concepts of FDA and functional classification trees and their ensembles.
Section 3 presents augmented functional classifiers and suggests possible tools for models' explainability.
Section 4 provides an application for the ECG200 dataset. 
Section 5 presents a simulation study with six scenarios. 
Finally, Section 6 concludes the paper, discussing findings, conclusions, and future research directions.

\section{Preliminaries}
\label{sec2:matmed}

\subsection{Functional Data Analysis (FDA)}
\label{sec21:matmed}

In FDA, the core idea is to treat data functions as distinct entities, but in practical scenarios, functional data is often encountered as a series of discrete data points. This implies that the original function, denoted by $z = f(x)$, is transformed into a collection of discrete observations represented by $T$ pairs $(x_j, z_j)$, where $x_j$ signifies the points where the function is assessed, and $z_j$ represents the corresponding function values at those points \citep{Ramsay2005, Ramsay2002, Ramsay2009, Ferraty2003}.

We define a functional variable $X$ as a random variable with values in a functional space $\Xi$. Consequently, a functional data set is a sample {$x_1, x_2, ..., x_N$} drawn from the functional variable $X$ \citep{Ramsay2005, Ramsay2002, Ramsay2009, Ferraty2003}.
Focusing our attention to the case of a Hilbert space with a metric $d(\cdot,\cdot)$ associated with a norm so that $d(x_1(t), x_2 (t)) = \|x_1(t) - x_2(t)\|$, and where the norm $\|\cdot \|$ is associated with an inner product $\langle \cdot,\cdot \rangle$ so that $\|x(t)\|=\langle x(t),x(t) \rangle^{1/2}$, we can obtain as a specific case the space $\mathcal{L}_2$ of real square-integrable functions defined on $\tau$ by $\langle x_1(t),x_2(t) \rangle=\int_{\tau} x_1(t)x_2(t)\text{d}t$, where $\tau$ is a Lebesgue measurable set on $T$. Therefore, by considering the specific case of $\mathcal{L}_2$, a basis function system consists of a set of linearly independent functions $\phi_j(t)$ that span the space $\mathcal{L}_2$ \citep{Ramsay2005}.

The initial step in FDA involves transforming the observed values $z_{i1}, z_{i2}, ..., z_{iT}$ for each unit $i=1,2,...,N$ into a functional form. The prevailing method for estimating functional data is basis approximation. Various basis systems can be employed depending on the characteristics of the curves. 
A common approach is to represent functions using a finite set of basis functions in a fixed basis system. This can be mathematically expressed as:

\begin{equation}
x_i(t) \approx \sum_{s=1}^S c_{is}\phi_s(t),
\label{smoothfun}
\end{equation}

\noindent where $c_i = (c_{i1}, ..., c_{iS})^T$ represents the vector of coefficients that define the linear combination, $\phi_s(t)$ is the $s$-th basis function, and $S$ is a finite subset of functions used to approximate the full basis expansion.

A trending methodology involves the utilization of a data-driven basis, with the Functional Principal Components (FPCs) decomposition standing out as a prominent technique. This approach effectively reduces dimensionality while concurrently preserving essential information from the original dataset \citep{Ramsay2005, Aguilera2013, Febrero2012}. In this context, the approximation of functional data can be expressed as follows:

\begin{equation} x_i(t) \approx \sum_{k=1}^{K} \nu_{ik}\xi_k(t), \label{fpca} \end{equation}
where $K$ is the number of FPCs, $\nu_{ik}$ represents the score of the generic FPC $\xi_k$ for the generic function $x_i$ ($i=1,2,...,N$).

By reducing the representation to the first \( p \) Functional Principal Components (FPCs), we can estimate the sample curves. The explained variance is calculated as \(\sum_{k=1}^p \lambda_k\), where \(\lambda_k\) represents the variance associated with the \( k \)-th FPC. The FPCs are constructed so that the variance explained by each \( k \)-th FPC decreases as \( k \) increases.

In this setting, under the assumption that the observed curves are centered with a null sample mean, the score for the $i$-th curve and $k$-th Functional Principal Component (FPCs) is determined by:

\begin{equation}
    \nu_{ik} = \int_\tau x_i(t)\xi_k(t)\text{d}t \;\;\; i=1,\cdots, N,
\end{equation}

 where $\xi_k(t)$ is the weight function\footnote{The weight function $\xi_k$ is obtained by maximizing the variance, solving the following optimization problem:
$Max_{\xi_k} Var\Bigl[\int_\tau X(t)\xi_k(t),\text{d}t \Bigr], $
subject to the constraints:
$||\xi_k||^2 = \int \xi_k(t)^2 \text{d}t = 1, $
and
$\int \xi_k(t)\xi_l(t), \text{d}t=0 \; \; \; \; \text{for} \; \; l\neq k. $}.

\subsection{Functional Classification Trees (FCTs)}

In the realm of functional classification, the objective is to forecast the class or label $Y$ for an observation $X$ within a separable metric space $(\Xi, d)$. Consequently, our methodology is tailored for functional data represented as ${y_i, x_i(t)}$, where $x_i(t)$ is a predictor curve defined for $t \in T$, and $y_i$ denotes the scalar response observed at sample $i = 1, ..., N$. The classification of a novel observation $x$ from $X$ involves the creation of a mapping $f:\Xi \longrightarrow \lbrace 0, 1, ... , U \rbrace$, referred to as a ``\textit{classifier},'' which assigns $x$ to its predicted label. The probability of error is quantified by $P \lbrace f(X) \neq Y \rbrace$.

The continuous domain $T$ can take on various forms, such as time, space, or other parameters. In this context, the primary focus is on the temporal domain, although the methodology can be seamlessly extended to incorporate other parameters. Theoretically, the response could manifest as either categorical or numerical, leading to classification or regression challenges, respectively. Nevertheless, the present investigation centers on a specific scenario: a scalar-on-function classification problem. 

Traditional decision trees represent a supervised learning technique that predicts response values by acquiring decision rules derived from features. Extensive information on decision trees can be found in various prior works \citep{Hyafil1976, Quinlan1986, Hastie2009}. Building upon this foundation, decision trees can be extended to the FDA framework \citep[see e.g.][]{Maturo2022SIM, maturo2023supervised}. This extension involves leveraging the coefficients of a basis representation as novel features for training a functional classifier. The latter methodology suggested tools to enhance the interpretability of the so-called ``\textit{Functional Classification Tree''} (FCT). 
The starting idea is that FCTs can be based both on fixed and data-driven basis systems. In the instance of a fixed basis system, such as the one presented in Equation \ref{smoothfun}, the matrix of features is determined by:

\begin{equation}
\mathbf{C}=
\begin{pmatrix}
    c_{11} & \dots  & c_{1S}  \\
    \vdots & \ddots   &   \vdots \\
    c_{N1} &  \dots   & c_{NS}
\end{pmatrix},
\label{featuresspline}
\end{equation}
where the generic element $c_{is}$ is the coefficient of the $i$-th curve ($i = 1,...,N$) relative to the $s$-th  ($s = 1,...,S$) basis function $\phi_s(t)$. As a natural consequence, $ \bf{c}_{i}$ is the vector containing the $i$-th statistical unit's characteristics. This strategy guarantees a straightforward application of the classifier. Indeed,  by focusing on b-splines as an example, owing to their total count and order, it becomes feasible to employ them to represent both the functional training set and the functional test set. Nevertheless, using a fixed basis system is frequently restrictive from an interpretative standpoint. This is due to the fact that basis functions are not derived from the data itself. Consequently, their significance in the time domain fails to align with the criteria of maximizing explained variability and finding a reliable interpretation of the classification rules.

On the contrary, in the case of a data-driven basis system like that in Equation \ref{fpca}, the features' matrix is given by: 

\begin{equation}
\mathbf{V}=
\begin{pmatrix}
    \nu_{11} & \dots  &  \nu_{1K}  \\
    \vdots & \ddots   &   \vdots \\
     \nu_{N1} &  \dots   &  \nu_{NK} 
\end{pmatrix},
\label{featuresfpca}
\end{equation}

\noindent where $\nu_{ik}$ is the score of the $i$-th curve ($i = 1,...,N$) relative to the $k$-th functional principal component  $\xi_{k}$ ($k = 1,...,K$). Feeding FCTs via FPCs is very useful in terms of dimensionality reduction, interpretability, and accuracy, as shown in \citep{Maturo2022SIM}.

\section{Augmented Functional Tree-Based Classifiers and their Transparency}

\subsection{Augmented Functional Classification Trees (AFCTs)}
\label{sec22:augtree}

The approach proposed by Maturo and Verde \citep{Maturo2022SIM} neglects the potential benefits of certain FDA techniques because it is limited to the original functions without considering the possibility of observing other functional characteristics. This research seeks to extend the latter approach and enhance classifier performance by leveraging alternative functional tools. Introducing the concept of  ``\textit{Augmented Functional Classification Tree}'' (AFCT), this study employs sequential derivatives of the original functional data to generate new features for feeding AFCTs. Essentially, it involves viewing functions from various angles (different dimensions given by the derivatives' orders), analogous to using a magnifying glass on curves, to uncover additional features that can significantly improve classification performance compared to what the original functions can capture.

Let the functional derivative of order $r$ for the $i$-th curve be represented by a fixed basis system (e.g. b-splines) as:
\begin{equation}
x_i^{(r)}(t)=\sum_{j=1}^S c_{i j }^{(r)} \phi_j^{(r)}(t) \quad j=1, \ldots, S
\label{augmentedfeaturesspline}
\end{equation}

\noindent where $c_{i j }^{(r)}$ is the coefficient of the $i$-th curve, $j$-th b-spline, and $r$-th derivative order;
$\phi_j^{(r)}(t)$ is the $r$-th derivative of the $j$-th basis function. 

Emphasized by Ramsay and Silverman \citep{Ramsay2005}, the selection of the basis system plays a crucial role in estimating derivatives. It is essential to ensure that the chosen basis for representing the object can accommodate the order of the derivative to be calculated. In the case of b-spline bases, this implies that the spline's order must be at least one higher than the order of the derivative under consideration. In this specific research, we concentrate on a b-spline basis of fourth order. The highest degree of derivative adopted in this study is the second; consequently, there are no problems estimating the basis functions' derivatives. In this work, the estimate is made using the \textit{deriv.fd} function in the \textit{fda} $\mathrm{R}$ package \citep{Ramsay2005}.

The fundamental idea of this research is that the feature matrix introduced in Equation \ref{featuresspline} can be expanded into a kind of matrix composed of blocks of coefficients derived from functional representations of various orders of derivatives. Following this approach, the feature matrix for training the so-called AFCT becomes as follows:

\begin{equation}
\mathbf{C}^{(r)}=
\begin{pmatrix}
    c_{11}^{(0)} & \dots  & c_{1S}^{(0)} & c_{11}^{(1)} & \dots  & c_{1S}^{(1)} & \dots & c_{11}^{(r)} & \dots  & c_{1S}^{(r)} \\
    \vdots & \ddots & \vdots & \vdots & \ddots & \vdots & \vdots & \vdots & \ddots & \vdots \\
    c_{N1}^{(0)} & \dots & c_{NS}^{(0)} & c_{N1}^{(1)} & \dots & c_{NS}^{(1)} & \dots & c_{N1}^{(r)} & \dots & c_{NS}^{(r)}
\end{pmatrix}
\label{extendedfeaturesspline}
\end{equation}

\noindent where $\mathbf{C}^{(r)}$ indicates that the features used leverage representations up to the $r$-th derivative. Naturally, focusing attention solely on features derived from a representation based on original functions, velocity and acceleration of curves, the feature matrix or order two is given by:

\begin{equation}
\mathbf{C}^{(2)}=
\begin{pmatrix}
    c_{11}^{(0)} & \dots  & c_{1S}^{(0)} & c_{11}^{(1)} & \dots  & c_{1S}^{(1)} & c_{11}^{(2)} & \dots  & c_{1S}^{(2)} \\
    \vdots & \ddots & \vdots & \vdots & \ddots & \vdots & \vdots & \vdots & \ddots & \vdots \\
    c_{N1}^{(0)} & \dots & c_{NS}^{(0)} & c_{N1}^{(1)} & \dots & c_{NS}^{(1)} & c_{N1}^{(2)} & \dots & c_{NS}^{(2)}.
\end{pmatrix}
\label{extendedfeaturesspline2}
\end{equation}

The functional derivatives are in a different space spanned by diverse basis systems. Therefore, when using a multivariate approach that combines the coefficients of b-splines from different derivatives, it is important to proceed with caution. This is because the coefficients from these different derivatives can vary significantly in magnitude and variability.
For this reason, using this approach with some statistical methods would require a standardization of scores. However, this is not necessary for the context of FCTs because, in classification trees, the standardization of features is not necessary; effectively, the different magnitudes and variability of the scores do not influence the split based on impurity reduction computed considering the categorical outcome.

Following the same criteria and starting from a representation like that of Equation \ref{featuresfpca}, we can also adapt the feature matrix illustrated in \ref{featuresfpca}.  
Let Equation \ref{featuresfpca} be the FPCs approximation of the original dataset.  
The functional derivatives can be rewritten as a decomposition in an orthonormal basis by maximizing the variance of $x^{(r)}(t)$:

\begin{equation}
\hat{x}^{(r)}_{i}(t)=\sum_{i=1}^{K}\nu^{(r)}_{ik}\xi^{(r)}_{k}(t)
\label{fpca}
\end{equation}

\noindent where $\nu^{(r)}_{ik}$ is the score for the $i$-th statistical unit for the $k$-th FPC $\xi^{(r)}_{k}(t)$ of order $r$, i.e. the derivative of order $r$  for the generic FPC $\xi_k(t)$ ($i=1,2,...,N$).

The generic matrix of data-driven augmented features becomes the following:

\begin{equation}
\mathbf{V}^{(r)}=
\begin{pmatrix}
    \nu_{11}^{(0)} & \dots  & \nu_{1S}^{(0)} & \nu_{11}^{(1)} & \dots  & \nu_{1S}^{(1)} & \dots & \nu_{11}^{(r)} & \dots  & \nu_{1S}^{(r)} \\
    \vdots & \ddots & \vdots & \vdots & \ddots & \vdots & \vdots & \vdots & \ddots & \vdots \\
    \nu_{N1}^{(0)} & \dots & \nu_{NS}^{(0)} & \nu_{N1}^{(1)} & \dots & \nu_{NS}^{(1)} & \dots & \nu_{N1}^{(r)} & \dots & \nu_{NS}^{(r)}
\end{pmatrix}
\label{extendedfeaturesFPC}
\end{equation}

In practical applications, it may be appropriate to stop at the second derivative, obtaining as the data-driven matrix to train the AFCT using the set of augmented features in Equation \ref{extendedfeaturesFPC2}:

\begin{equation}
\mathbf{V}^{(2)}=
\begin{pmatrix}
    \nu_{11}^{(0)} & \dots  & \nu_{1S}^{(0)} & \nu_{11}^{(1)} & \dots  & \nu_{1S}^{(1)} & \nu_{11}^{(2)} & \dots  & \nu_{1S}^{(2)} \\
    \vdots & \ddots & \vdots & \vdots & \ddots & \vdots & \vdots & \vdots & \ddots & \vdots \\
    \nu_{N1}^{(0)} & \dots & \nu_{NS}^{(0)} & \nu_{N1}^{(1)} & \dots & \nu_{NS}^{(1)} & \nu_{N1}^{(2)} & \dots & \nu_{NS}^{(2)}.
\end{pmatrix}
\label{extendedfeaturesFPC2}
\end{equation}

The algorithm commences with the entire functional dataset, comprising the scores derived from the decomposition of successive derivatives illustrated in Equation \ref{extendedfeaturesFPC2}. The procedure continues iterating until reaching terminal leaves. At each step, the algorithm determines the optimal binary partition based on the selected cost criterion, such as the Gini index or Shannon-Wiener index\footnote{The Gini index measures heterogeneity for categorical variables, with lower values indicating greater homogeneity within a node. Conversely, the Shannon-Wiener entropy index quantifies heterogeneity, with higher values signifying greater diversity within a node. The Gini index and Shannon-Wiener entropy index can be calculated as follows:
\begin{equation}
G=\sum_m^F 1-f_{rm}^2,
\label{gini}
\end{equation}
\begin{equation}
E=\sum_m^F f_{rm} \cdot \text{ln} f_{rm},
\label{shannon}
\end{equation} where $f_{rm}$ represents the proportion of training observations in the $r$-th region that are from the $m$-th class.}\citep{Hastie2009, rpart}, as described in \citep{Hastie2009} and \citep{rpart}. The initial classifier is an extensive AFCT, which is then refined through cost-complexity pruning to achieve an appropriate balance between complexity and accuracy providing the pruned AFCT.

\subsection{Augmented Functional Classification Trees' Interpretability}
\label{sec23:interpraugtree}

One of the main challenges with Augmented Functional Classification Trees (AFCTs) is understanding the classification rules in the context of functional data. In AFCTs, as explained in Section \ref{sec22:augtree}, the coefficients \(\nu^{(r)}_{ik}\) from successive linear combinations are used as features for predicting the response. However, interpreting the split rules in AFCTs differs from traditional decision trees. This is because the split values for \(\nu^{(r)}_{ik}\) must be understood based on the specific part of the time domain that the corresponding Functional Principal Component (FPC) represents, as well as the derivative order. Therefore, by examining both the split value of \(\nu^{(r)}_{ik}\) and the plot of \(\xi^{(r)}_{k}(t)\), one can gain a better understanding of the rules that separate the functional dataset at each node.

Let the subscript "0" in $\nu_{0k}^{(r)}$ signify that the identified threshold for the score of order $r$ associated with a specific Functional Principal Component $\xi_k(t)^{(r)}$ is set as a constant value to divide the set of curves into two subgroups. For instance, in the context of the initial split rule, namely the separation rule for the root node, $\nu_{0k_0}^{(r)}$ represents the threshold value linked to FPC $k_0^{(r)}$. Consequently, all curves meeting the condition $\nu_{ik_0}^{(r)} < \nu_{0k_0}^{(r)}$ constitute one subgroup, while the remaining functions, those satisfying the condition $\nu_{ik_0}^{(r)} \geq \nu_{0k_0}^{(r)}$, form the other subset.

To better understand the split rule in the functional context, we have to think about the theoretical separation curve generated by every split occurring in the AFCT. In a AFCT, each intermediate node produces a separation of the functional dataset into two functional subsets (children nodes). The curve that dictates the separation rule is given by:

\begin{equation}
\psi_z(t) = \sum_{\xi_k^{(r)}(t) \in \omega}\nu_{0k}^{(r)}\xi_{k}^{(r)}(t),
\label{FPSP}
\end{equation}

\noindent where $\omega=\{\xi_k^{(r)}(t)_{z_1},...,\xi_k^{(r)}(t)_{z_Z} \}$ is the set of the FPCs $\xi_k^{(r)}(t)$ selected in the classification rule path until the split of the $z$-th node ($z=1,..., Z$). The generic intermediate node that generates a separation is therefore indicated with $z$, and a total number of these intermediate nodes is identified with $Z$. In other words, each $\psi_z(t)$ can be associated with every intermediate node; by definition, $\psi_1(t)$ is the separation curve dictated to split the root node whereas the leaves (terminal nodes) do not contemplate a separation curve $\psi_z(t)$.

If, for instance, a intermediate node is obtained starting from the root node, first splitting based on the value 3.4 of the coefficient of the first derivative of the third FPC, and then splitting based on the value 2.1 of the coefficient of the second derivative of the first FPC, we will need to reconstruct this separation curve using Equation \ref{FPSP}. The curve reconstructed in this manner allows us to understand the reason for the separation in the time domain and thus from a functional perspective.

\subsection{Augmented Functional Random Forest (AFRF)}
\label{sec24:augrf}

In traditional decision trees, even slight variations in the data can result in highly diverse trees, leading to different classification rules and interpretations. A practical approach to mitigating this problem is constructing an ensemble of decision trees using bagging, as outlined by Hastie (2009)\citep{Hastie2009}.
Classical bagging, an extension of conventional decision trees proposed by Breiman (1996)\citep{Breiman1996}, aims to reduce the variance associated with a single classification tree. The fundamental concept involves generating multiple decision trees to formulate a final classification rule collectively; however, a limitation of bagging lies in the dominance of the most influential predictor across all trees. Consequently, the trees exhibit some degree of correlation, diminishing the strength of variance reduction. The adoption of the random forest can mitigate the latter issue.
The classical Random Forest (RF), introduced by Ho (1998)\citep{Ho1998}, stands out as a highly efficient machine learning algorithm and represents a specific case of bagging tailored for decision trees. The approach involves applying bagging to the data and employing bootstrap sampling for predictor variables at each decision split. This means that, during each step of the tree algorithm, a random subset of predictors is chosen as candidates for splitting from the entire set of predictors. This refinement enhances the traditional bagging method by producing a classifier less influenced by correlations among trees, preventing the dominance of a single discriminating variable across all trees.

A similar procedure can be extended to the case of functional data and lead to the Augmented Functional Random Forest (AFRF), i.e. a classification rule for augmented functional data via the FPCs scores used as features. Clearly, the same approach can be used with other bases, including b-splines. The difference from the classic case is that the basic weak classifiers are FCTs.
Let AFRF consist of $H$ AFCTs $\omega_h$, where $h=1,...,H$, with $H$ chosen to be a large number. The $h$-th AFCT $\omega_h$ is grown on a random subset of the training set, obtained from the original data $D = {(y_i , x_i(t)), i = 1,..., N}$ by drawing, with replacement, a bootstrap sample $D^\ast_h={(y_s^{(h)}, x_s^{(h)}(t)), s =1, ..., N}$ of the same size $N$ as the original data set.
The replacement of $x_s^{(h)}(t)$ can be seamlessly executed using the features matrix introduced in \ref{extendedfeaturesFPC2}. AFRF enhances the bagging process through a slight modification that decorrelates the AFCTs and reduces variance when averaging the AFCTs.

Each time a split in a single AFCT is considered, a random selection of $m$ of $\nu^{(r)}_{ik}$ is chosen as split candidates from the full set of the $K \cdot (r+1)$ FPCs. It follows that when $m<K \cdot (r+1)$, we have AFRFs, whereas when $m=K \cdot (r+1)$, we have an augmented functional bagging.
Following this approach, the AFCT into the forest, will be less correlated because the most important FPCs won't always be those features on the top of the AFCT determining the first important separation rules of the AFRF.
Every time a split in a single AFCT is considered, a random selection of $m$ $\nu^{(r)}_{ik}$ is chosen as split candidates from the complete set of $K \cdot (r+1)$ FPCs. It follows that when $m<K \cdot (r+1)$, we have AFRFs, whereas when $m=K \cdot (r+1)$, we have augmented functional bagging.
By adopting this approach, the AFCTs within the forest become less correlated because the most critical FPCs will only sometimes be those features positioned at the top of the AFCT determining the initial significant separation rules of the AFRF.
A helpful guideline to follow is to select the size of the subset of FPCs as a value of approximately $m \approx \sqrt{K \cdot (r+1)}$.
This means that during each split in the FCT, the algorithm will only consider some of the available augmented features.
On average, \(\frac{K \cdot (r+1)-m}{K \cdot (r+1)}\) of the splits will not even take some augmented features into account. This process helps AFRF to decorrelate the AFCTs, making the average less variable and thus more reliable.

The collection of curves labelled from $i = 1$ to $N$ that exist in the $h$-th bootstrap sample $D^\ast_h$ are referred to as the ``in-bag curves sample'' (IBCs). These curves are used to create a single $h$-th AFCT. The "out-of-bag curves sample" (OOBCs) comprises the remaining curves of statistical units that are not present in $D^\ast_h$.
We train $H$ AFCTs, each using its own bootstrapped functional training set, to get $\hat{f}^{*h}(x_i(t))$, which is the predicted class for curve $i$ using the $h$-th AFCT. We then combine all $H$ predictions of the AFCTs to obtain the final prediction for each curve $i$ using a "majority vote" system. For each curve, the predicted class is the most frequently occurring class among the $H$ predictions from the different AFCTs.

Let \( Y \) be a categorical random variable representing the class label, with \( U+1 \) possible classes, denoted by \( \delta \in \{0, 1, \dots, U\} \). For a given curve \( x_i(t) \), the goal is to estimate the probability \( P(y_i = \delta) \) that the curve belongs to class \( \delta \). This probability can be estimated by aggregating the predictions from an ensemble of \( H \) Augmented Functional Classification Trees (AFCTs). Specifically, the probability estimate is given by:

\begin{equation}
\hat{P}(y_i = \delta) = \frac{1}{H} \sum_{h=1}^{H} I\left( \hat{f}^{*h}(x_i(t)) = \delta \right),
\end{equation}

\noindent where \( I(\cdot) \) is the indicator function, defined as:

\begin{equation}
I\left( \hat{f}^{*h}(x_i(t)) = \delta \right) = 
\begin{cases} 
1 & \text{if } \hat{f}^{*h}(x_i(t)) = \delta, \\
0 & \text{otherwise}.
\end{cases}
\end{equation}

The final prediction of the Augmented Functional Random Forest (AFRF) for the curve \( x_i(t) \) is determined by the mode of the predicted classes across the ensemble. Formally, the AFRF prediction \( \hat{f}_{AFRF}(x_i(t)) \) is given by:

\begin{equation}
\hat{f}_{AFRF}(x_i(t)) = \underset{\delta \in \{0, 1, \dots, U\}}{\text{argmax}} \, \hat{P}(y_i = \delta).
\end{equation}

Alternatively, this can be expressed as:

\begin{equation}
\hat{f}_{AFRF}(x_i(t)) = \delta \quad \text{if and only if} \quad \hat{P}(y_i = \delta) \geq \hat{P}(y_i = \kappa) \quad \forall \kappa \in \{0, 1, \dots, U\}, \kappa \neq \delta.
\end{equation}

In this formalism, the AFRF assigns the \( i \)-th curve to the class \( \delta \) with the highest estimated probability, effectively taking the class predictions' mode across all ensemble trees.

In the case of AFRF, interpretability can not be considered since the method is based on utilizing a set of AFCTs. Each AFCT is characterized by the presence of coefficients derived from the use of successive derivatives of various FPCs of different order, which operate at different levels of the AFCTs with different split values. Moreover, these coefficients may not appear in a specific AFCT due to randomness of bootstrap. Therefore, discussing interpretability in this context is less meaningful than what is discussed in Section \ref{sec22:augtree}.
However, within the traditional random forest context, various explainability tools are available, and all tools documented in high-level non-functional literature can be easily extended to the functional domain following the outlined methodology. In this work, we propose an algorithm specifically adapted to address the inherent correlation present among our features. Given that the features in our model are constructed in a way that naturally introduces correlation—particularly when derived from functional data—we have developed a modified approach to ensure that these correlations are correctly accounted for in the analysis. This adaptation is presented in Section \ref{sec25:ciafpcs} and is crucial for maintaining the integrity of the model's predictions and enhancing the reliability of the classification process.

\subsection{Augmented Functional Random Forest Explainability: Conditional Permutation Importance for Augmented Functional Principal Components}
\label{sec25:ciafpcs}

In this section, we introduce and justify the development of novel feature importance metrics designed explicitly for AFRF. While effective in many contexts, conventional methods of measuring feature importance in Random Forests, such as Gini Importance (Mean Decrease in Impurity) and Permutation Importance (Mean Decrease in Accuracy) may only be somewhat appropriate for functional data that includes derivatives. This inadequacy arises primarily due to the inherent correlations among the scores of functional principal components derived from successive derivatives, which can introduce bias into classical importance measures.
The classical measures work under the assumption that the predictors are independent or only weakly correlated. However, in the case of augmented functional data, where features are derived from successive derivatives, these assumptions are violated. This violation can lead to overestimating the importance of certain FPCs, as their correlated nature artificially inflates their apparent contribution to reducing node impurities.
Therefore, it becomes necessary to develop feature importance metrics that can account for this correlation and provide a more accurate representation of each feature's contribution to the overall model performance.

In the literature, \citet{strobl2008conditional} suggested methods for conditioning based on groupings and extended the classical permutation approach by conditioning on the correlated features. Instead of permuting the values of a single feature across all observations, the proposal is to permute the values within groups of observations that share similar values in the correlated features. 
Our context is quite peculiar because it is evident that each FPC is inherently correlated with its first and second derivatives. Thus, standard permutation importance methods can also lead to misleading conclusions about relative importance. 
To address this issue, we propose a novel method, \textit{Conditional Permutation Importance for Augmented Functional Principal Components (CPIAFPCs)}, which calculates the importance of each FPC's score while conditioning on the scores of its corresponding dimensions.
This approach ensures that each score's importance reflects its unique contribution, independent of the influence of its correlated derivatives.

The proposed algorithm is designed to accurately measure FPCs importance in Augmented Functional Random Forests when dealing with functional data that includes derivatives. The process for calculating the Conditional Permutation Importance for Augmented Functional Principal Components (CPIAFPCs) is detailed in Algorithm \ref{alg}.

\begin{algorithm}[htbp]
\caption{Conditional Permutation Importance for Augmented Functional Principal Components (CPIAFPCs)}
\label{alg:CPIAFPCs}
\begin{algorithmic}[1]
    \State \textbf{Input:} 
    \begin{itemize}
        \item Matrix of augmented features $\mathbf{V}^{(2)}$, where the rows represent different observations $i = 1, \ldots, N$, and the columns represent the scores $\nu_{ik}^{(r)}$ for the Functional Principal Components (FPCs) of order $r = 0, 1, 2$ for each component $k = 1, \ldots, K$.
        \item Augmented Functional Random Forest (AFRF) model $\hat{f}_{AFRF}$ trained on $\mathbf{V}^{(2)}$.
    \end{itemize}
    \State \textbf{Grouping of Correlated Features:}
    \begin{itemize}
        \item Group together  features corresponding to the same principal component across different derivative orders. For each principal component $k$, form a group $G_k = \{\nu_{ik}^{(0)}, \nu_{ik}^{(1)}, \nu_{ik}^{(2)}\}$.
    \end{itemize}
    
    \State \textbf{Conditional Permutation:}
    \For{each group $G_k$ corresponding to the $k$-th principal component}
        \For{each feature $\nu_{ik}^{(r)}$ within the group $G_k$}
            \State \textbf{Within-Group Permutation:}
            \begin{itemize}
                \item Condition on the other features in the group $G_k$.
                \item Permute the values of $\nu_{ik}^{(r)}$, maintaining the correlated features constant.
            \end{itemize}
            \State \textbf{Model Prediction:}
            \begin{itemize}
                \item Use the permuted dataset to predict the response $\hat{y}_i^{\text{perm}}$ using the trained AFRF model $\hat{f}_{AFRF}$.
            \end{itemize}
            \State \textbf{Calculate Conditional Permutation Importance:}
            \begin{itemize}
                \item Compute the difference in prediction accuracy before and after permutation:
                \begin{equation}
                    \text{CPIAFPC}_{\nu_{ik}^{(r)}} = \frac{1}{N} \sum_{i=1}^{N} \left( \mathcal{L}(y_i, \hat{y}_i) - \mathcal{L}(y_i, \hat{y}_i^{\text{perm}}) \right)
                \end{equation}
                where $\mathcal{L}(y_i, \hat{y}_i)$ is the loss function measuring prediction accuracy, and $\hat{y}_i^{\text{perm}}$ are the predictions after permutation.
            \end{itemize}
        \EndFor
    \EndFor
    
    \State \textbf{Output:} 
    \begin{itemize}
        \item A set of importance scores $\text{CPIAFPC}_{\nu_{ik}^{(r)}}$ for each feature $\nu_{ik}^{(r)}$ in the dataset, reflecting the unique contribution of each feature to the model's performance.
    \end{itemize}
\end{algorithmic}
\label{alg}
\end{algorithm}

\section{Application to Electrocardiogram Data}
\label{sec3:appreal}

This section delves into the outcomes derived from applying AFCT and AFRF to an ECG dataset, providing comparisons with various functional classifiers. In particular, our methodology is tested on the ECG200 dataset accessible at https://www.timeseriesclassification.com/.
In 2001, R. Olszewski introduced the ECG200 dataset at Carnegie Mellon University as part of his work titled ``Generalized feature extraction for structural pattern recognition in time-series data'' \citep{ecg200dataset}. This dataset has been continuously used to evaluate new classifiers and currently holds the world record for classification accuracy, with the BOSS algorithm achieving 89.05\%.
The focus here is on using the original functions and their first and second derivatives. 

\begin{figure}[!htb]
\minipage{0.5\textwidth}
  \includegraphics[width=\linewidth]{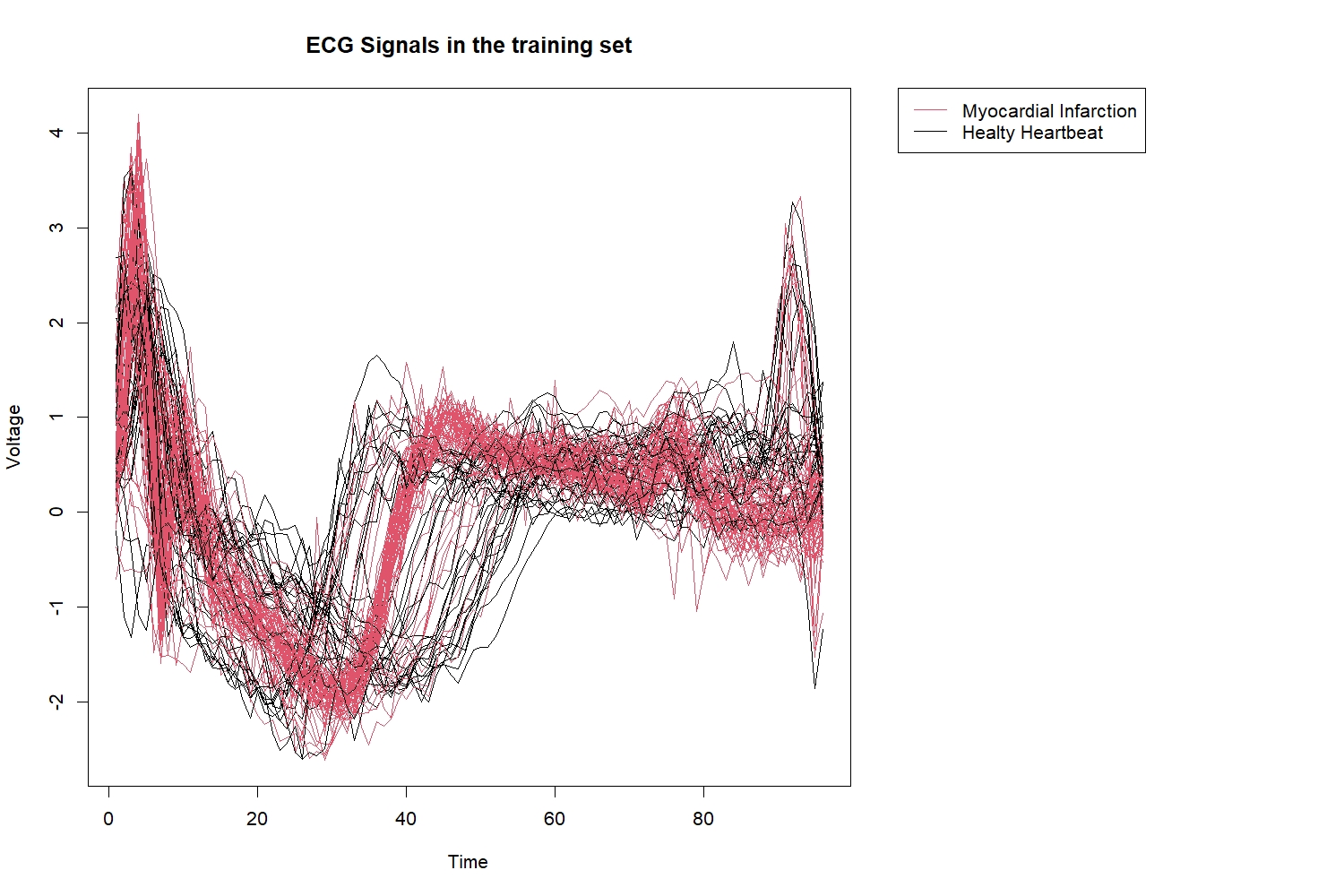}
  \caption{ECGs in the training set.}
  \label{grezzo1}
\endminipage\hfill
\minipage{0.5\textwidth}
  \includegraphics[width=\linewidth]{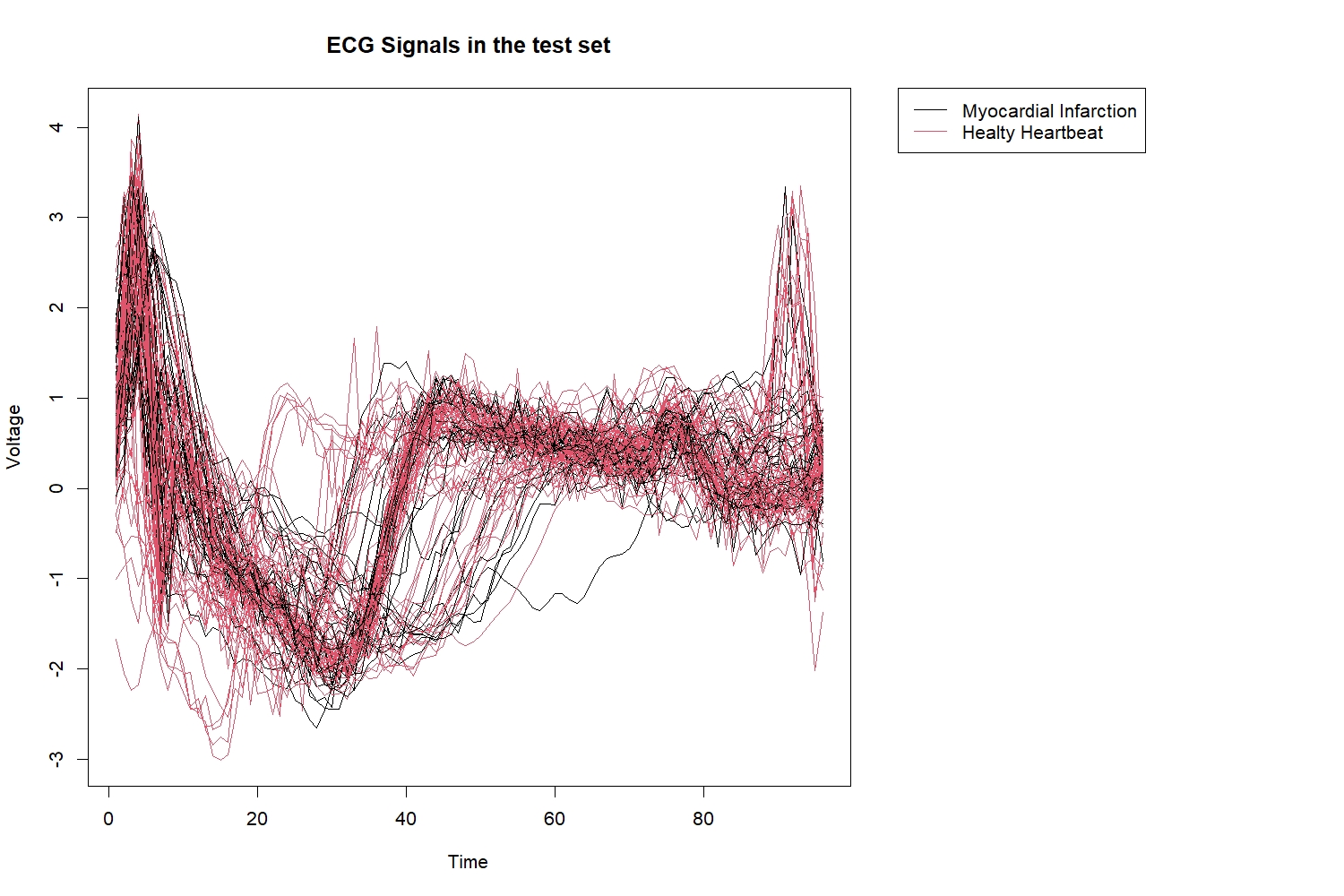}
  \caption{ECGs in the test set.}
  \label{grezzo2}
\endminipage
\end{figure}

\begin{figure}[!htb]
\minipage{0.5\textwidth}
  \includegraphics[width=\linewidth]{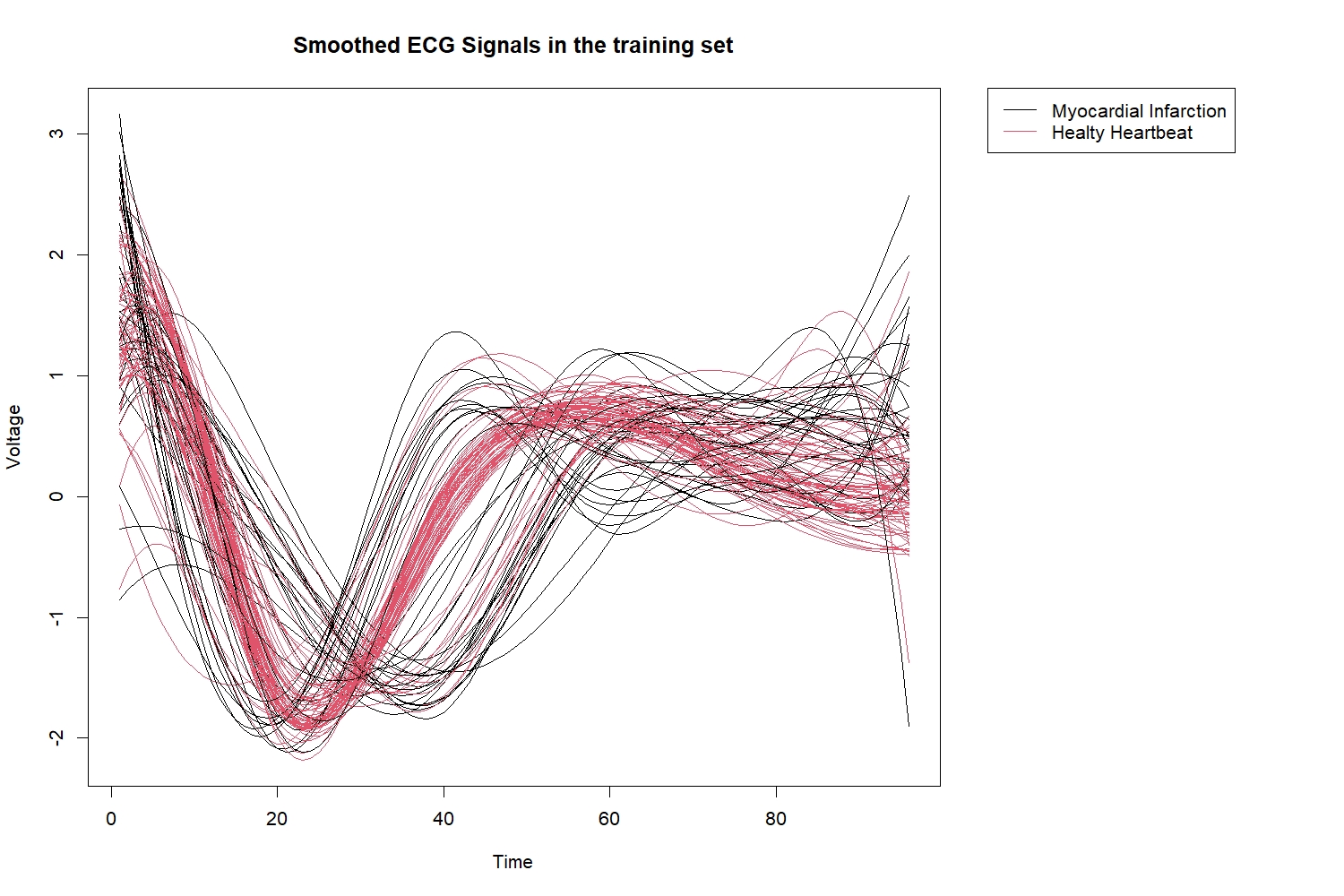}
  \caption{Smoothed ECGs in the training set.} 
  \label{smoo1}
\endminipage\hfill
\minipage{0.5\textwidth}
  \includegraphics[width=\linewidth]{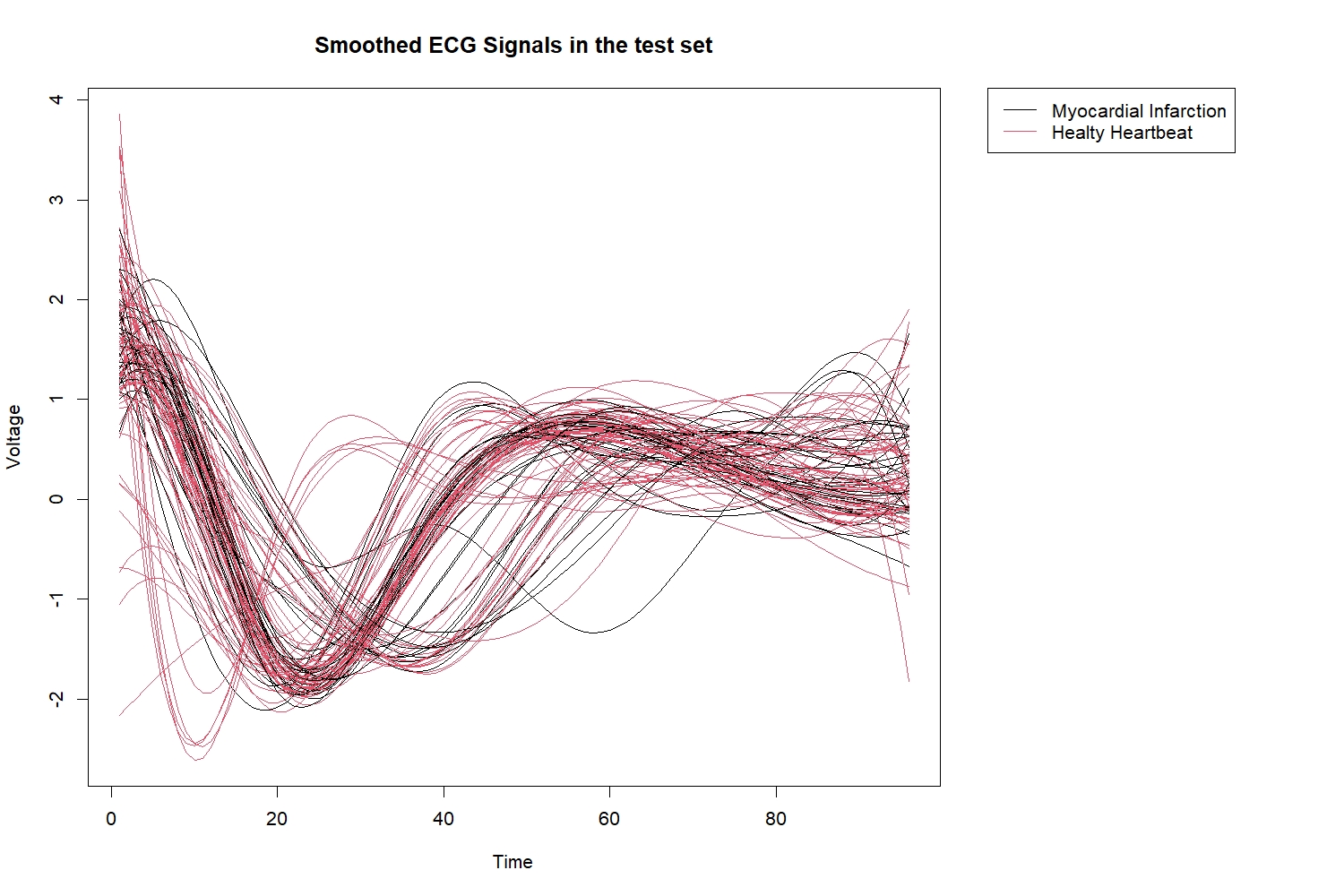}
  \caption{Smoothed ECGs in the test set.}
  \label{smoo2}
\endminipage
\end{figure}

Figures \ref{grezzo1} and \ref{grezzo2} show the smoothed versions of the original signals in the training and test sets. Each series represents the electrical activity that is recorded during a single heartbeat. The two categories are a normal heartbeat (represented by the colour red) and a Myocardial Infarction (represented by the colour black). In our training set, we have a total of 100 signals, and in the test set, we also have 100 signals.
Equation \ref{smoothfun} was used to compute smoothed versions of the original signals in the training and test sets. These smoothed versions are presented in Figures \ref{smoo1} and \ref{smoo2}.

The objective is to develop a classifier that can forecast a patient's health status based on their ECG readings. It's worth noting that by utilizing cubic splines with an order of four, we can maintain the continuity of the splines' first and second derivatives at the knots.

\begin{figure}[htpb]
\centering
\includegraphics[width=8cm]{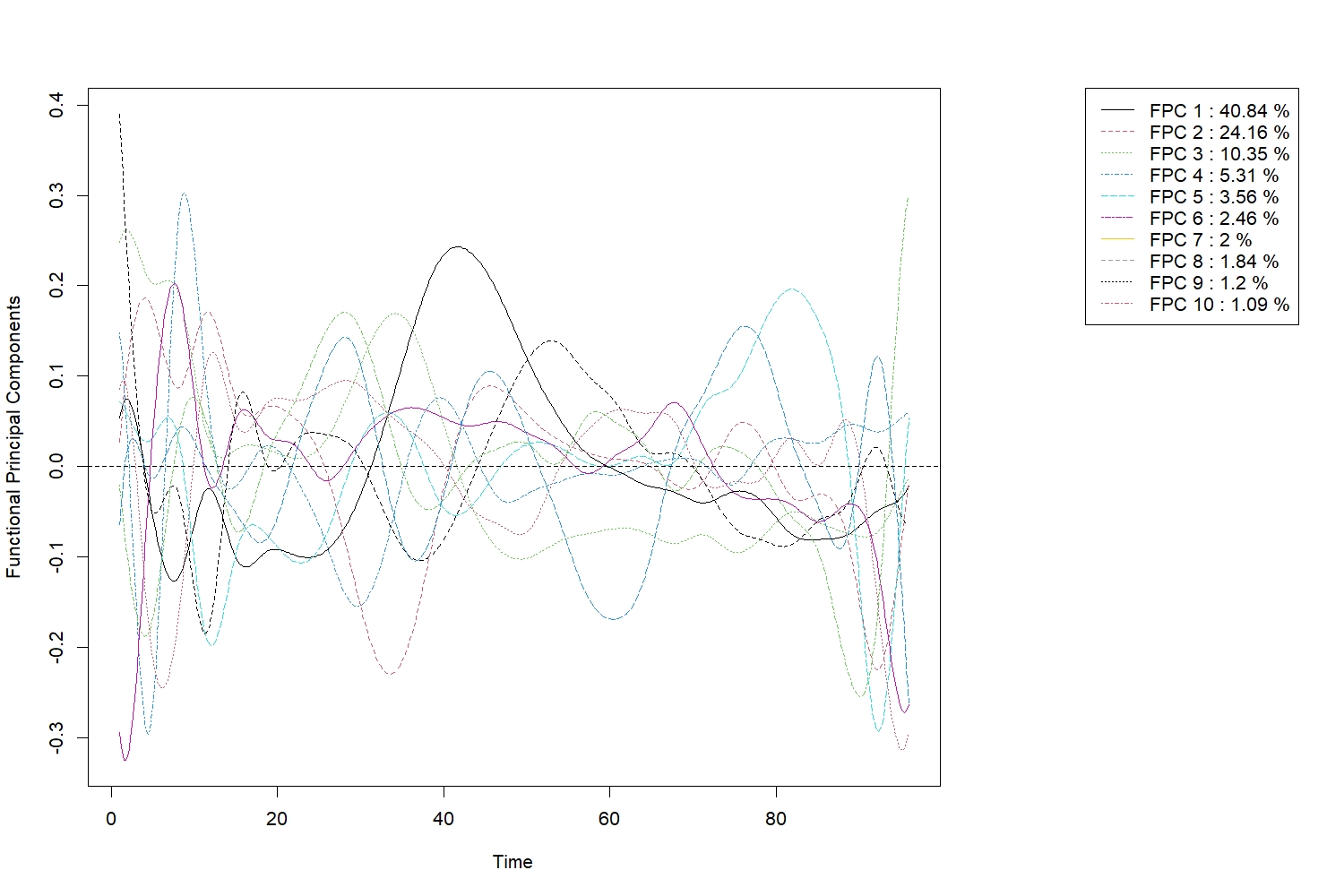}
\caption{First Ten Functional Principal Components of the original curves.}
\label{FPCs}
\end{figure}

\begin{figure}[!htb]
\minipage{0.5\textwidth}
  \includegraphics[width=\linewidth]{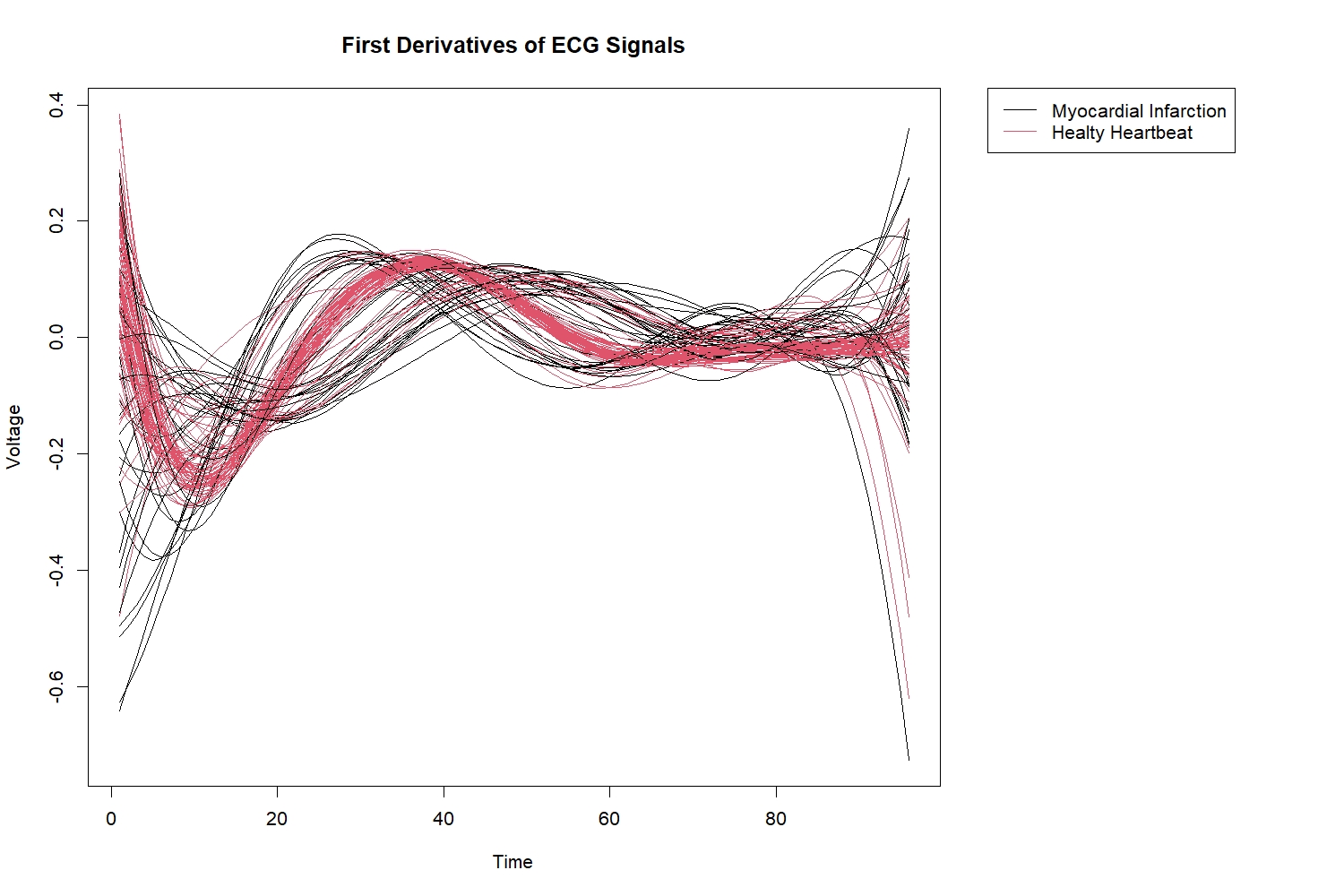}
\caption{First derivatives of the original functions.}
\label{d1}
\endminipage\hfill
\minipage{0.5\textwidth}
  \includegraphics[width=\linewidth]{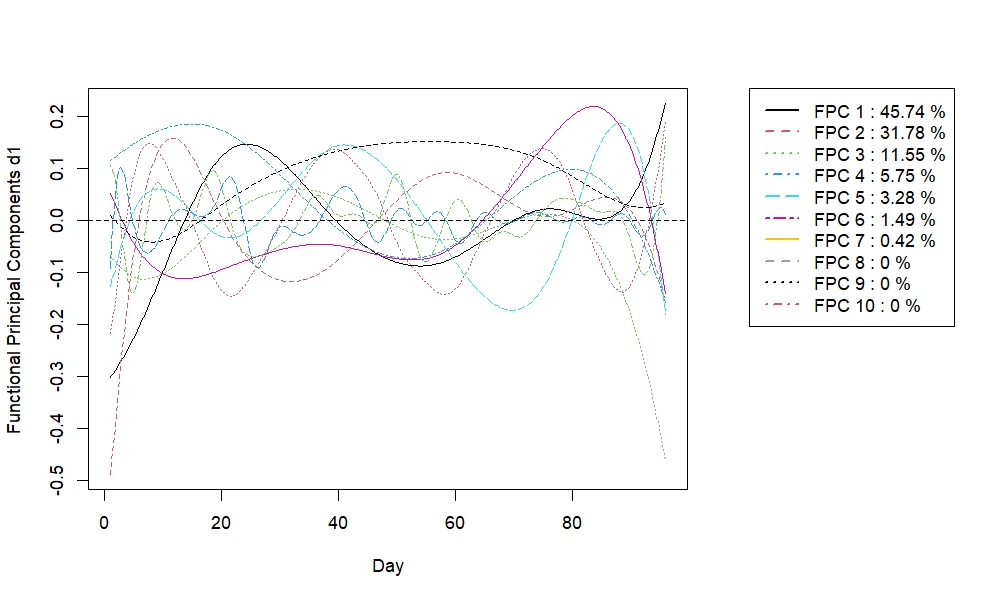}
\caption{FPCs' first derivatives.}
  \label{d1FPCs}
\endminipage
\end{figure}

\begin{figure}[!htb]
\minipage{0.5\textwidth}
  \includegraphics[width=\linewidth]{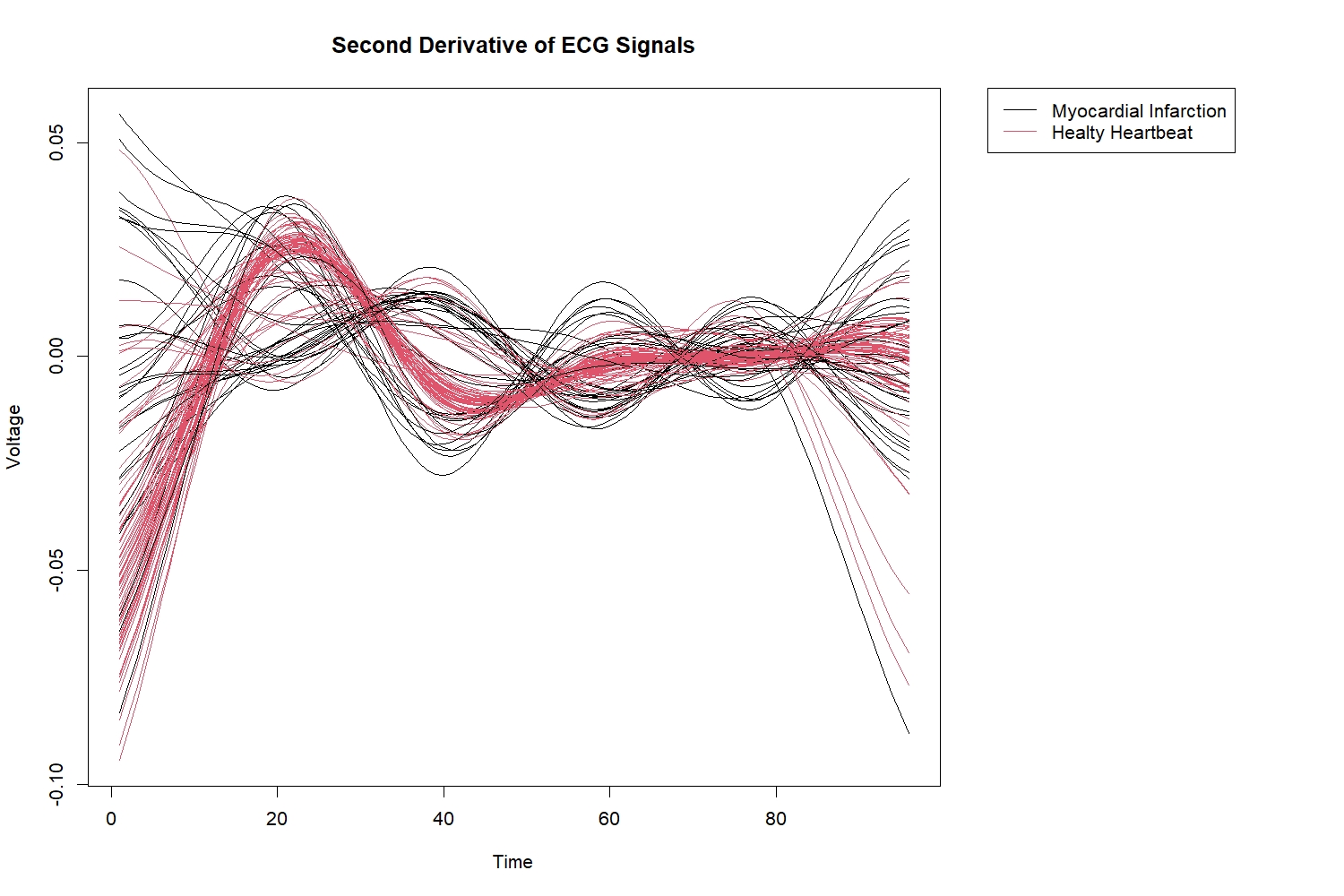}
\caption{Second derivatives of the original functions.}
  \label{d2}
\endminipage\hfill
\minipage{0.5\textwidth}
  \includegraphics[width=\linewidth]{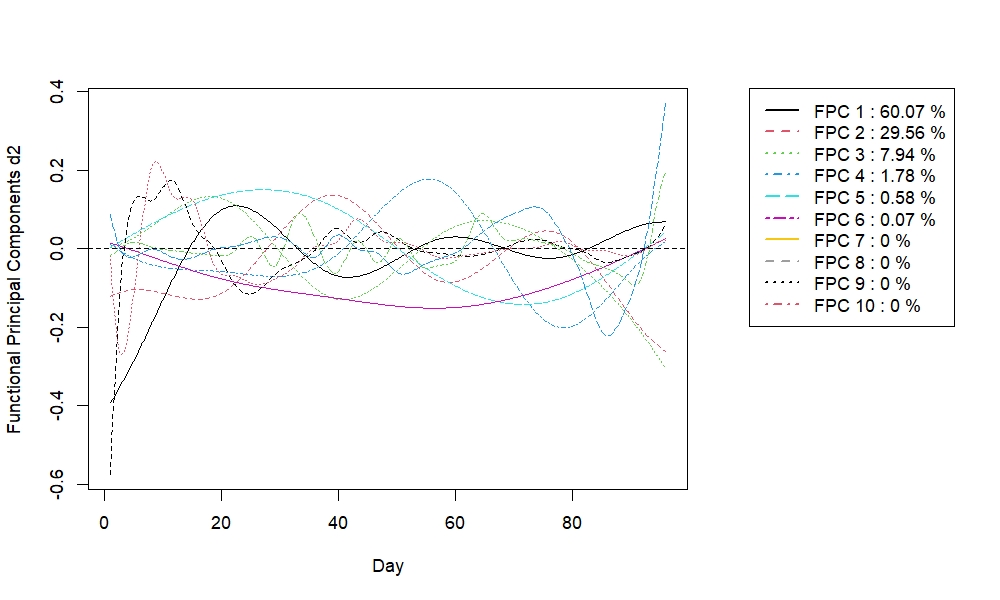}
\caption{FPCs' second derivatives.}
    \label{d2FPCs}
\endminipage
\end{figure}

Figure \ref{FPCs} displays the original curves' first ten functional principal components. Figures \ref{d1} and \ref{d1FPCs} show the first derivatives of the original functions and the FPCs' first derivatives respectively. Similarly, Figures \ref{d2} and \ref{d2FPCs} display the second derivatives of the original functions and the FPCs' second derivatives. The scores used in Figures \ref{FPCs}, \ref{d1FPCs}, and \ref{d2FPCs} are fed into the AFCT presented subsequently. It is important to note that in this framework, the traditional methods of selecting the number of FPCs are not helpful. This is because FPCs that explain little variability can often be crucial in distinguishing between the outcome classes and are, therefore, essential to constructing the AFCT. In fact, the first FPC, which typically captures most of the variability, is rarely critical in AFCTs. As Figure \ref{FPCs} illustrates, each FPC explains different parts of the time domain and can be useful in distinguishing between the outcome classes in the context of AFCTs.

\begin{figure}[htpb]
\centering
\includegraphics[width=10cm]{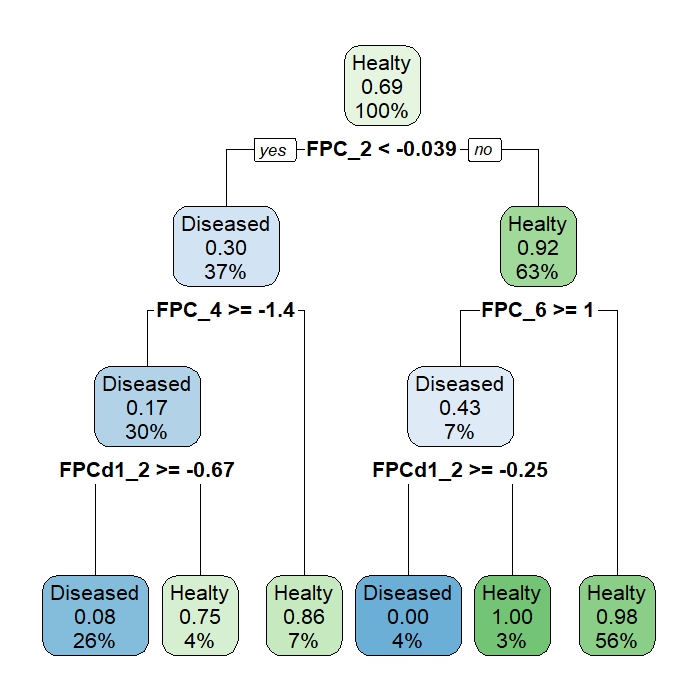}
\caption{Augmented (pruned) Functional Classification Tree (AFCT) via Successive FPCs' Derivatives (AFCT). Accuracy of 94\% on the training set (apparent error) and 83\% on the test set.}
\label{AFCTpicture}
\end{figure}

Non-pruned AFCT is typically not useful for practical purposes. If the AFCT overfits the data, it will perform poorly when classifying different datasets. For this reason, the pruning phase is essential. This phase is based on finding an optimal balance between complexity and accuracy. Cost-complexity pruning via cross-validation is performed using the R package ``\textit{rpart}''. The pruned AFCT is shown in Figure \ref{AFCTpicture}. It was built using FPCs scores computed using Equation \ref{fpca} as features via the \textit{rpart.plot} R package. The cut on a specific value of an FPC score determines the split of a node. Among all the possible FPCs and splitting score values of all the derivatives' orders, the one that maximizes the decrease of impurities of the node is chosen.

The AFCT Classifier has an accuracy of 95\% on the training set (apparent error) and 83\% on the test set, which shows an improvement compared to an FCT without derivatives.

\begin{figure}[htpb]
\centering
\includegraphics[width=12cm]{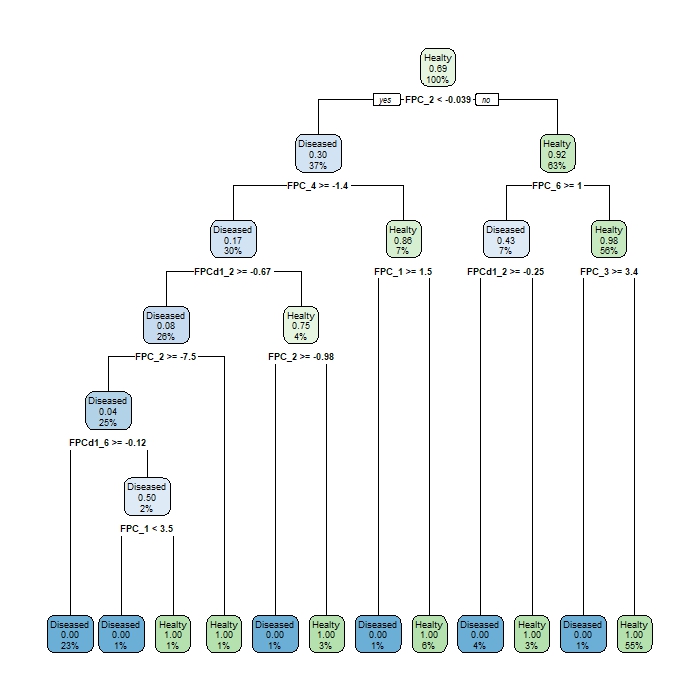}
\caption{Augmented (non-pruned) Functional Classification Tree (AFCT) via Successive FPCs' Derivatives (AFCT).}
\label{AFCTpicturenonpruned}
\end{figure}

\begin{figure}[htpb]
\centering
\includegraphics[width=8cm]{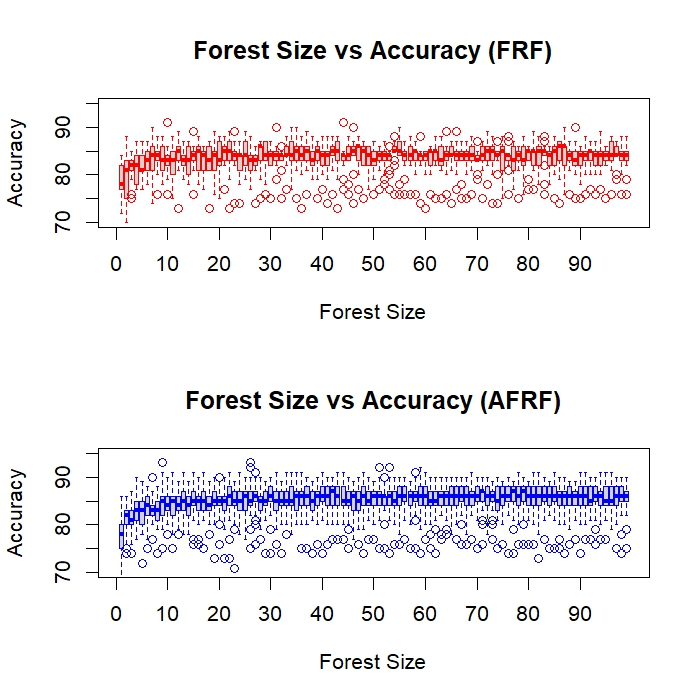}
\caption{The distributions of the accuracy for FRF and AFRF.}
\label{comparemedian}
\end{figure}

\begin{figure}[htpb]
\centering
\includegraphics[width=8cm]{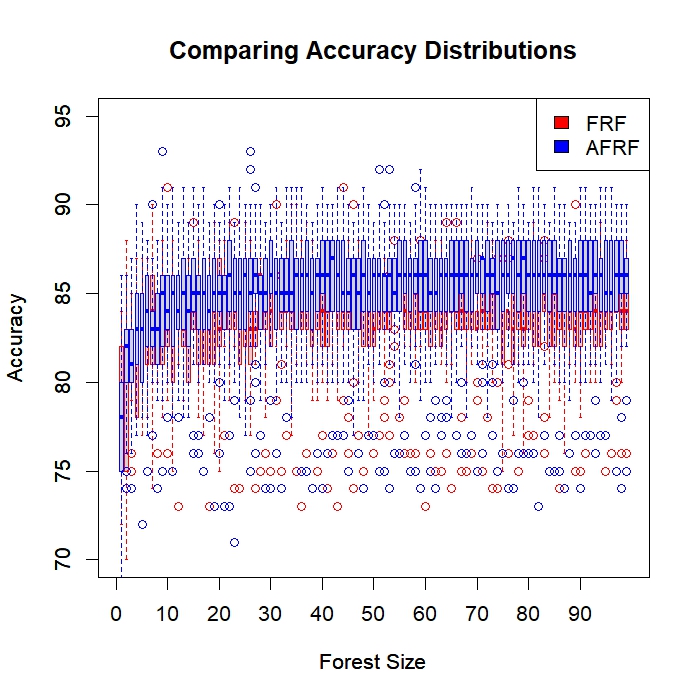}
\caption{Comparing Accuracy Distributions: FRF VS AFRF.}
\label{comparemediansovrapposte}
\end{figure}

We tested the AFRF algorithm on different forest sizes and numbers of FPCs to measure its accuracy. 
As a result, we have a distribution of values related to accuracy for each value of the forest size rather than a single value. This circumstance makes the results more robust because it significantly limits the impact of chance. We compared the results of the new algorithm with those of FRF without augmentation, always relying on using the test set.
Figure \ref{comparemedian} presents the results of the FRF and AFRF approaches. It can be observed that FRF exhibits peaks of maximum accuracy at 91\%, while AFRF shows peaks at 93\%. To better compare the accuracy distributions, we propose Figure \ref{comparemediansovrapposte}, where we overlap the boxplots and can easily appreciate that AFRF is consistently superior to FRF.

\begin{figure}[htpb]
\centering
\includegraphics[width=8cm]{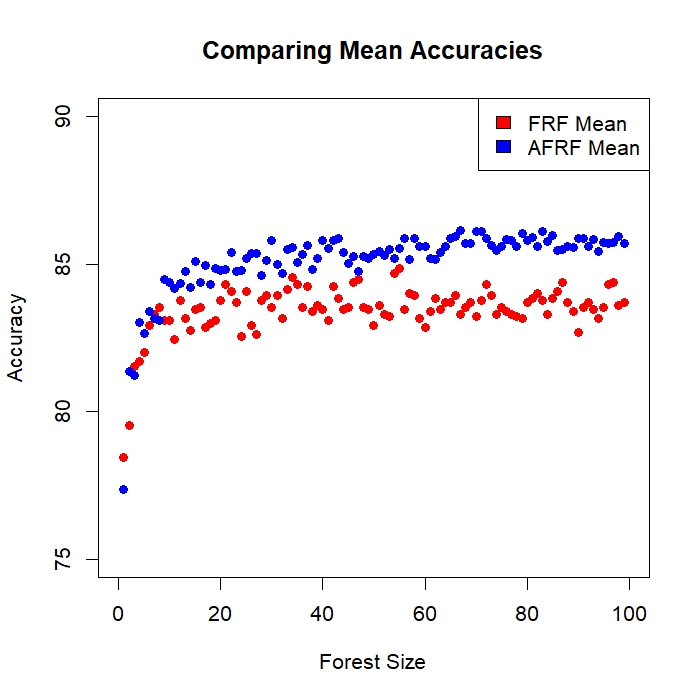}
\caption{Comparing Mean Accuracies: FRF VS AFRF.}
\label{compareboxplotresultsmedie}
\end{figure}

Further comparison is presented in Figure \ref{compareboxplotresultsmedie}, where we illustrate the mean accuracies instead of considering medians and quartiles. It is quite evident that AFRF is significantly superior, and this difference increases with the growth of the forest size and then stabilizes. A simple calculation of the means of the two algorithms (for all tree sizes and all possible numbers of FCPs) highlights that FRF has an overall mean accuracy of 83.43\%, while AFRF has an overall mean accuracy of 85.10\%. We stress that the previous world record for this dataset was set by the BOSS algorithm at 89.05\%.

To compare AFRF, based on FPCs, with the most recent and widely used methods for classifying functional data, we present the results of various approaches implemented in the fda.usc R package \citep{Febrero2012}. Table \ref{ECG2004} emphasizes that the other methods fail to achieve the performance of the AFRF classifier.
The functional K-NN classifier achieves a test set accuracy of 91.00\% with 3 or 5 nearest neighbors. Lastly, employing functional depth classifiers in Table \ref{ECG2005} , the highest accuracy on the test set is 81.00\% with the depth measure "RP." In summary, the AFRF classifier based on FPCs demonstrates compelling results and proves to be the most accurate among the tested methods for this dataset.

\begin{table}[ht]
\centering
\begin{tabularx}{\textwidth}{XXX}
  \hline
NNs & Accuracy\\ 
  \hline
1 & 89.00 \\ 
  3  & 91.00 \\ 
  5  & 91.00 \\ 
  7  & 88.00 \\ 
  9  & 86.00 \\ 
  11  & 88.00 \\ 
  13  & 85.00 \\ 
  15  & 82.00 \\ 
  17  & 82.00 \\ 
  19  & 80.00 \\ 
   \hline
\end{tabularx}
\caption{Functional classification using the k-nn classifier of the R package fda.usc. The best accuracy on the test set is 91.00\% with 3 or 5 nearest neighbours.}
\label{ECG2004}
\end{table}

\begin{table}[ht]
\centering
\begin{tabularx}{\textwidth}{XXX}
\toprule
Functional Depth & Accuracy \\
\midrule
RP & 81.00 \\
mode & 79.00 \\
RT & 43.00 \\
FM & 74.00 \\
RPD & 80.00 \\
\bottomrule
\end{tabularx}
\caption[Functional classification using depth classifiers]{Functional classification using depth classifiers of the R package fda.usc. The best accuracy on the test set is 81\% with the depth measure ``RP''. depth.RP computes the Random Projection depth (see Cuevas et al. 2007). depth.mode implements the modal depth (see Cuevas et al 2007). depth.RT implements the Random Tukey depth (see Cuesta-Albertos and Nieto-Reyes 2008). depth.FM computes the integration of an univariate depth along the axis x (see Fraiman and Muniz 2001). It is also known as Integrated Depth. depth.RPD implements a depth measure based on random projections possibly using several derivatives (see Cuevas et al. 2007).}
\label{ECG2005}
\end{table}

\begin{figure}[htbp]
    \centering
    \includegraphics[width=\textwidth]{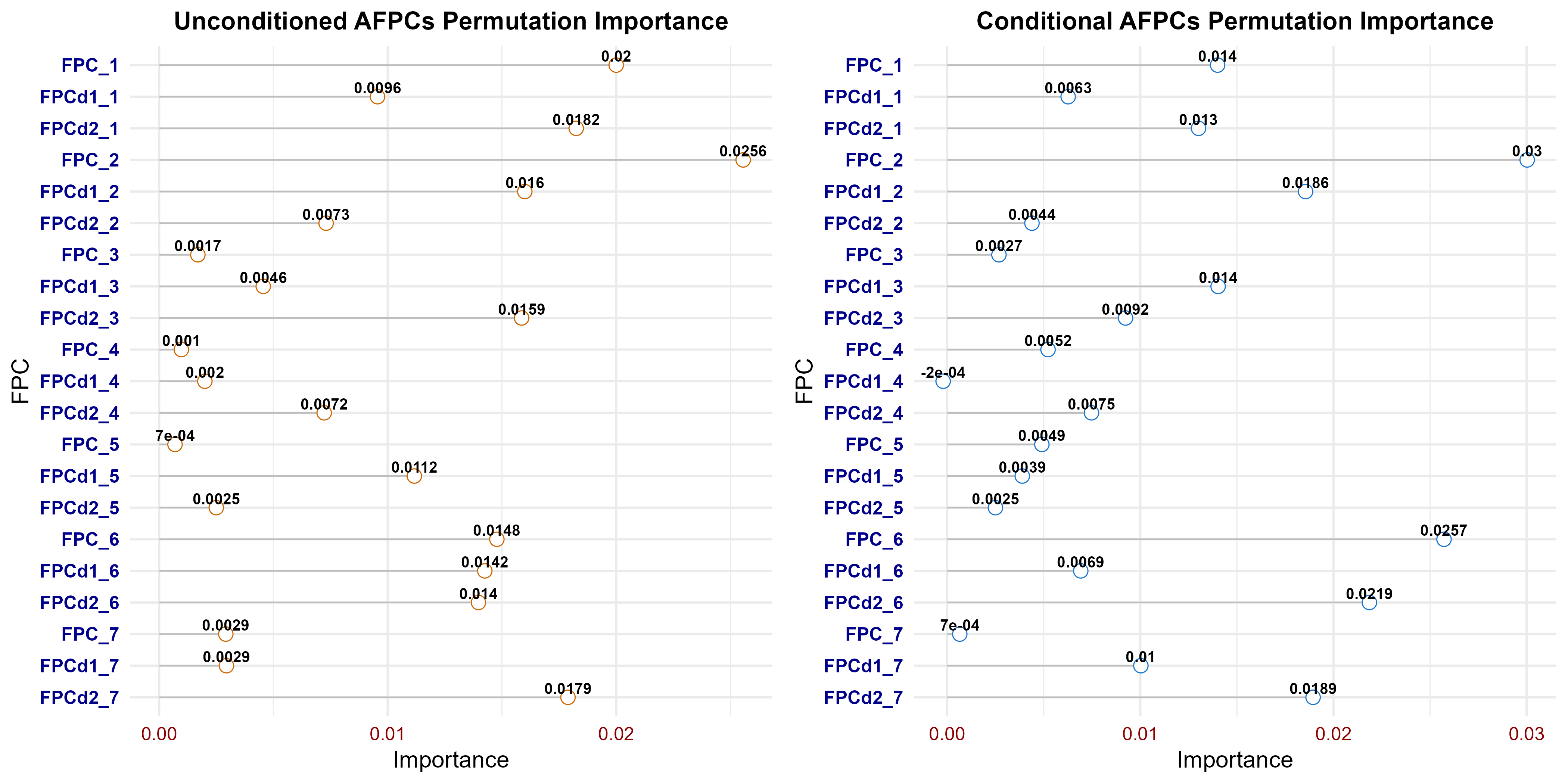}
    \caption{Comparison of Conditional vs. Unconditional Permutation Importance for Augmented Functional Principal Components (AFPCs). }
    \label{fig:importance_comparison}
\end{figure}

The Conditional Permutation Importance for Augmented
Functional Principal Components is displayed in Figure~\ref{fig:importance_comparison}.
We compare the Permutation Importance for Augmented
Functional Principal Components under two scenarios.
The left panel displays the Unconditional Importance, which measures the importance of each AFPC without considering the influence of other variables. The right panel shows the Conditional Importance, where the importance of each AFPC is evaluated while accounting for the effects of other scores of the corresponding dimensions. This side-by-side comparison reveals how the relevance of each AFPC changes when conditioning is applied. For instance, AFPCs like $FPCd\_6$ exhibit a substantial increase in conditional importance, suggesting they gain relevance when other variables are controlled. Conversely, some components, such as $FPC\_1$, demonstrate higher unconditional importance, indicating their overall influence without the conditioning effects. This comparison is essential for identifying which AFPCs are robust across different conditions and which are more sensitive to including other FPCs.

\begin{figure}[htbp]
    \centering
    \includegraphics[width=0.8\textwidth]{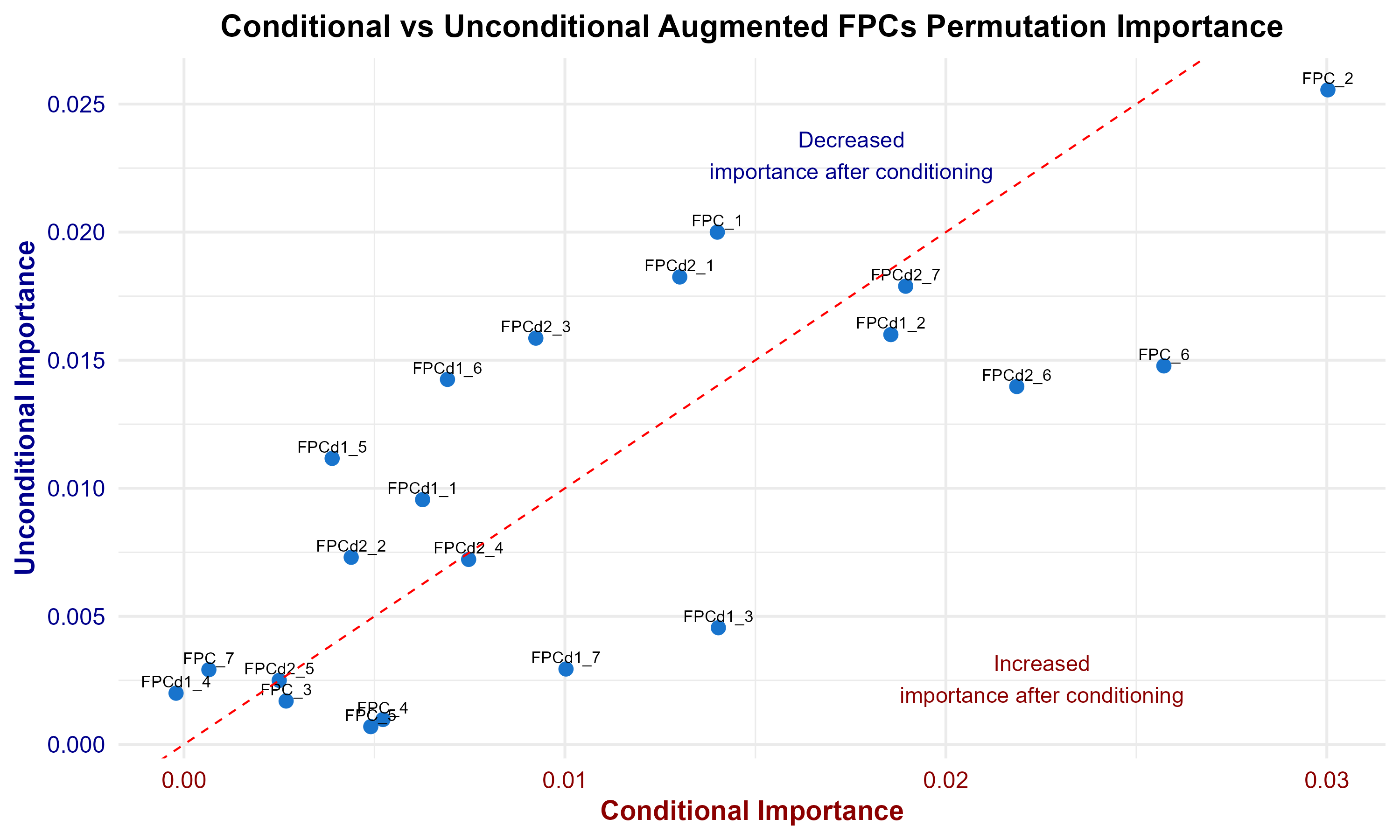}
    \caption{Scatter plot of Conditional vs Unconditional Importance Changes in Augmented FPCs. }
    \label{conditional_vs_unconditional_importance}
\end{figure}

Figure~\ref{conditional_vs_unconditional_importance} presents a scatter plot that compares the conditional and unconditional permutation importance of AFPCs. 
The scatter plot illustrates the relationship between conditional importance (x-axis) and unconditional importance (y-axis). The red dashed line represents the bisector, with equal importance values. AFPCs above this line (e.g., FPC\_1) have decreased in importance after conditioning, indicating that their significance is reduced when the effects of other variables are controlled. Conversely, AFPCs below the line (e.g., FPC\_2) have increased in importance, highlighting that they become more relevant when conditioned on other factors. This analysis helps to identify which AFPCs are more stable across different conditions and which ones are sensitive to changes in the analytical context.

\section{Simulation study}
\label{sec4:simu}

To demonstrate the effectiveness of the AFPCs classifier, we customize and refine various models previously examined by researchers such as \citep{Cuevas_2007, Preda_2007, maturo2023supervised} to generate functional data with distinctive shapes. We present six scenarios in which we generate 100 functions from each group, forming training and test sets of different sizes according to the number of balanced classes (200 curves when three classes are present, 300 curves when dealing with three groups, etc.).
In all instances, data are equally divided into training and test sets, and each curve spans a domain of 50-time observations.
For each simulation, we conduct comparisons between FRF and AFRF. The diverse scenarios are generated through the following simulations.

\begin{figure}[htpb]
\centering
\includegraphics[width=\textwidth]{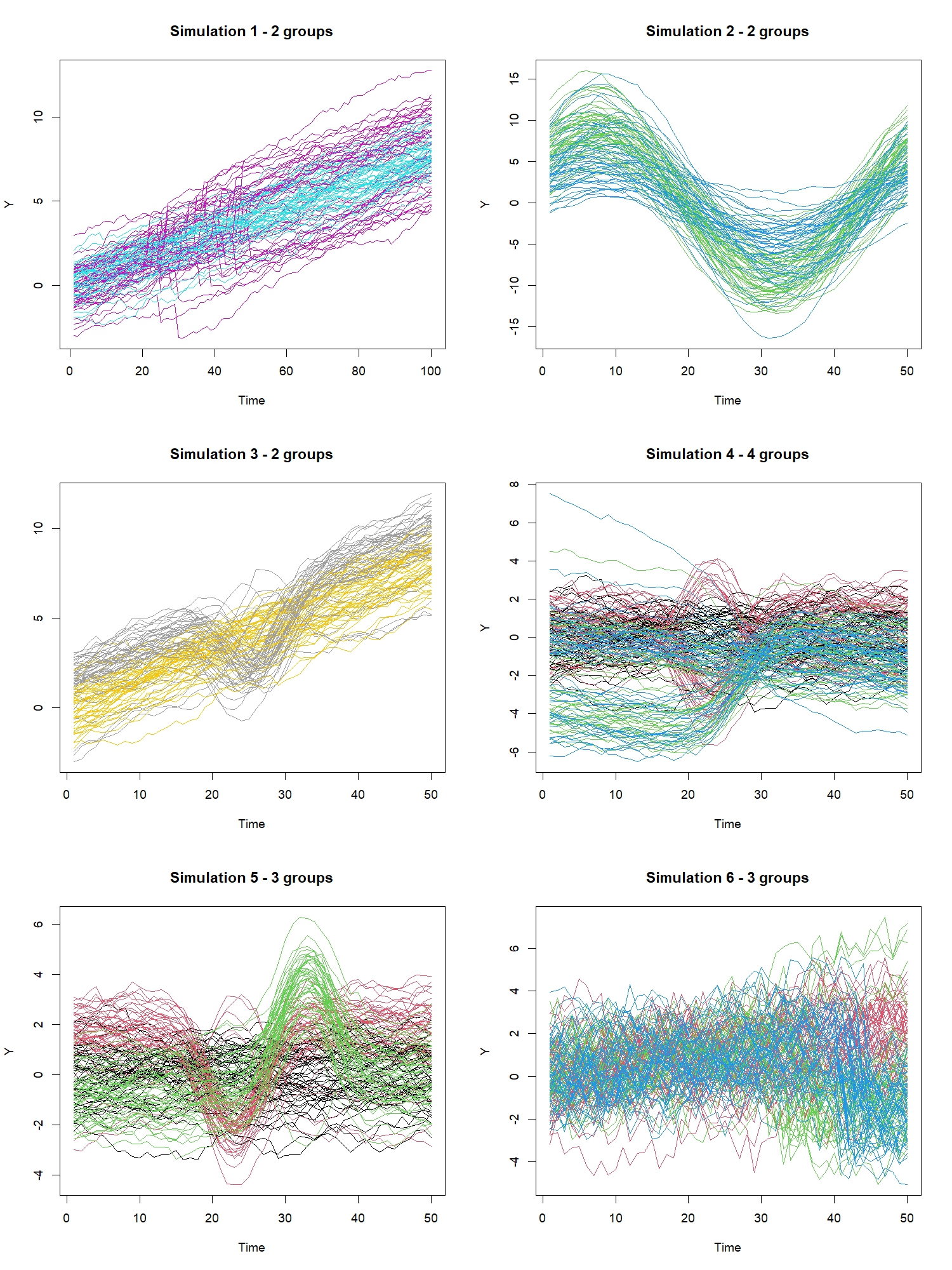}
\caption{Simulated scenarios of functional data with two, three, and four classes to predict.}
\label{simulazioni}
\end{figure}

Figure \ref{simulazioni} illustrates the six datasets generated using the model described above.

\textbf{Simulation 1}. Group 1 is modeled by the equation $X_{i}(t)=\mu t+e_{i}(t)$, while Group 2 is represented by the equation $X_{i}(t)=\mu t+q k_{i} I_{T_{i} \leq t}+e_{i}(t)$, where $t$ ranges from 0 to 1. The term $e_{i}(t)$ denotes a Gaussian process characterized by a zero mean and covariance defined as $\gamma(s, t)=\alpha \exp \{-\beta|t-s|^{\nu}\}$. Here, $k_{i}$ takes values from the set ${-1,1}$ with equal probability, and $I$ is an indicator function. Furthermore, $\boldsymbol{q}$ serves as a constant determining the extent to which the curves in Group 2 deviate from the overall trend observed in Group 1. The variable $T_{i}$ is uniformly distributed within the interval $[a, b] \subset [0,1]$. The simulation enables the generation of two distinct groups with variations in their amplitudes limited to a specific segment of the time domain. For this simulation, we set the parameters: $\mu = 8$, $q = 2$, $a=0.2$, $b=0.5$, $\alpha=1$, $\beta=1$, and $\nu=1$.\\[\baselineskip]

\textbf{Simulation 2}. In this simulation, we employ two distinct functional data-generating models to create two groups, primarily distinguished by their amplitudes. The primary model takes the form $X_{i}(t)=a_{1 i} \sin \pi+a_{2 i} \cos \pi+e_{i}(t)$. To generate Group 2, we adopt the model $X_{i}(t)=\left(b_{1 i} \sin \pi+b_{2 i} \cos \pi\right)\left(1-u_{i}\right)+\left(c_{1 i} \sin \pi+c_{2 i} \cos \pi\right) u_{i}+e_{i}(t)$, where $t$ varies within the interval $[0,1]$, and $\pi$ is confined to $[0,2 \pi]$. The coefficients $a_{1 i}$ and $a_{2 i}$ follow a uniform distribution in the interval $\left[a_{1}, a_{2}\right]$, while $b_{1 i}$ and $b_{2 i}$ follow a uniform distribution in $\left[b_{1}, b_{2}\right]$. Additionally, $c_{1 i}$ and $c_{2 i}$ adhere to a uniform distribution within $\left[c_{1}, c_{2}\right]$. The binary variable $u_{i}$ follows a Bernoulli distribution, and $e_{i}(t)$ represents a Gaussian process characterized by a zero mean and covariance function given by $\gamma(s, t)=\alpha \exp \{ -\beta|t-s|^{\nu}\}$. Figure \ref{simulazioni} visually displays the simulated data, with parameters set as follows: $a_{1 i}=2$, $a_{2 i}=10$, $b_{1 i}=1.5$, $b_{2 i}=11.5$, $c_{1 i}=1$, $c_{2 i}=4$, $\alpha=2$, $\beta=0.5$, and $\nu=1$.\\[\baselineskip]

\textbf{Simulation 3}. In this simulation, we explore two distinct functional data-generating models to create two groups that exhibit slight differences both in magnitude and shape, specifically within a designated segment of the time domain. Group 1 is generated using the model $X_{i}(t)=\mu t+e_{i}(t)$. In contrast, Group 2 is produced by the model $X_{i}(t)=\mu t+(-1)^{u} \cdot q+(-1)^{(1-u)}\left(\frac{1}{\sqrt{r \pi}}\right) \exp \left(-z(t-v)^{w}\right)+e_{i}(t)$, where $t \in [0,1]$, $e_{i}(t)$ is a Gaussian process with a zero mean and covariance function defined as $\gamma(s, t)=\alpha \exp \{-\beta|t-s|^{\nu}\}$, $u$ follows a Bernoulli distribution with $P(u=1)=0.5$, and $\boldsymbol{q}$, $\boldsymbol{r}$, $z$, $\boldsymbol{w}$ are constants. Additionally, $v$ follows a Uniform distribution in the interval $[a, b]$. The parameters defining the two groups are set as follows: $\mu = 8$, $q = 1.8$, $a=0.45$, $b=0.55$, $\alpha=1$, $\beta=1$, $\nu=1$, $r = 0.02$, $z=90$, and $w=2$. This choice of parameters ensures that the groups differ in both amplitude and shape, contributing to a nuanced variation within a specific time frame.\\[\baselineskip]

\textbf{Simulation 4}. This simulation is based on the same model introduced for simulation three but with the difference that four different classes are created. Naturally, the parameters can be set differently. Two parameter configurations are generated to obtain four different groups. The first configuration is as follows: $\mu = 0$, $q = 1$, $a=0.45$, $b=0.45$, $\alpha=1.3$, $\beta=1.2$, $\nu=1$, $r = 0.02$, $z=90$, and $w=2$. The second configuration is as follows:
$\mu = -2$, $q = 1.8$, $a=0.15$, $b=0.15$, $\alpha=0.8$, $\beta=0.8$, $\nu=1$, $r = 0.01$, $z=90$, and $w=5$. \\[\baselineskip]

\textbf{Simulation 5}. This simulation is based on the same procedure introduced for simulation four but with the difference that three different classes are created. The first configuration is as follows: $\mu = 0$, $q = 1.8$, $a=0.45$, $b=0.45$, $\alpha=1$, $\beta=1$, $\nu=1$, $r = 0.02$, $z=90$, and $w=2$. The second configuration is as follows: $\mu = 1$, $q = 0.8$, $a=0.65$, $b=0.65$, $\alpha=1$, $\beta=1$, $\nu=1$, $r = 0.02$, $z=90$, and $w=2$.\\[\baselineskip] 

\textbf{Simulation 6}. This simulation follows the methodology introduced in simulation one, with the distinction that it involves the creation of three distinct classes. The first configuration is specified as follows: $\mu = 2$, $q = 3$, $a=0.6$, $b=0.75$, $\alpha=2$, $\beta=1$, and $\nu=0.5$. The second configuration is characterized by $\mu = 2$, $q = 3$, $a=0.8$, $b=0.9$, $\alpha=2$, $\beta=1$, and $\nu=0.5$.\\[\baselineskip]

\begin{figure}[htpb]
\centering
\includegraphics[width=\textwidth]{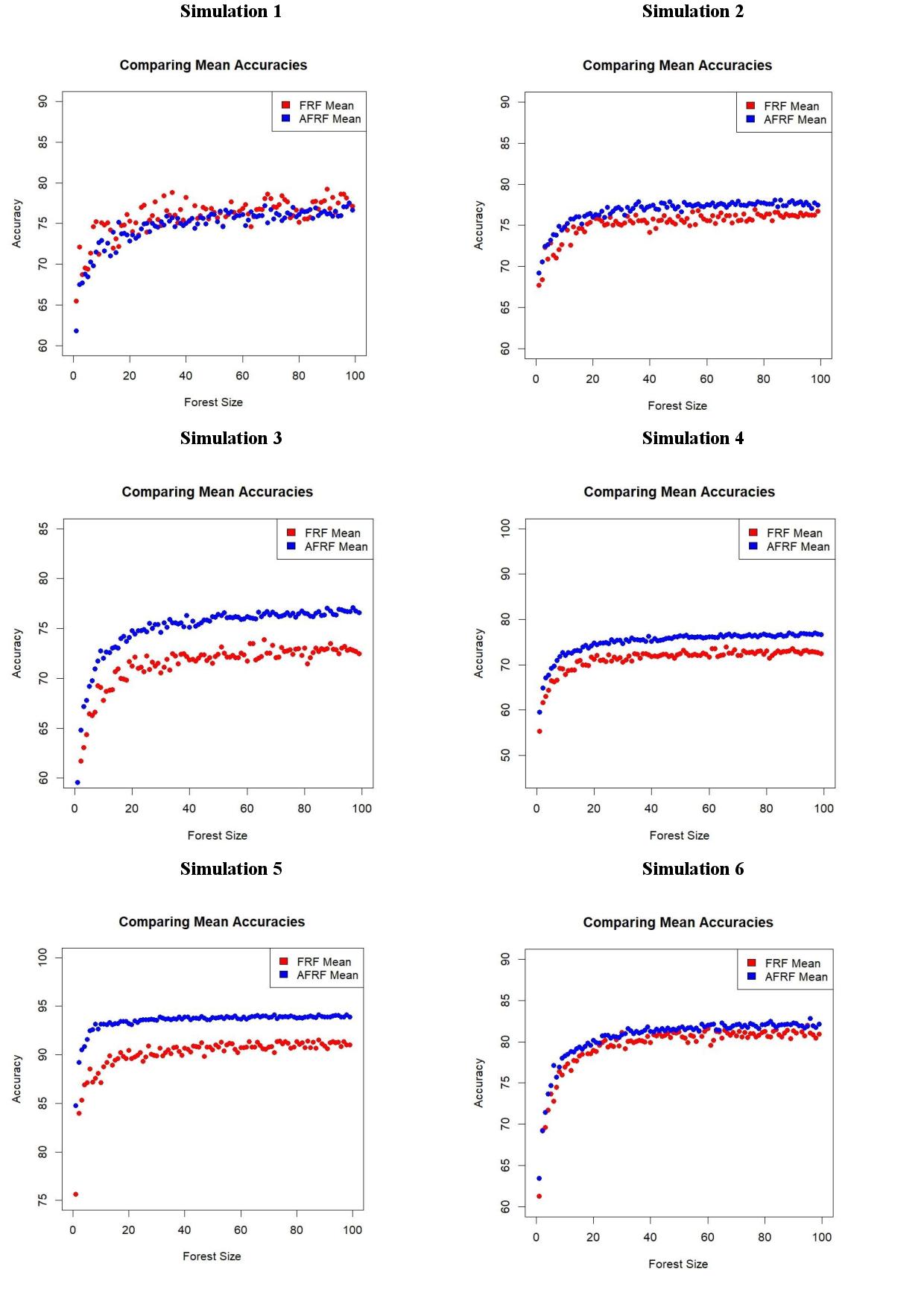}
\caption{Results of the simulated scenarios and comparison between FRF and AFRF in terms of mean accuracy.}
\label{Risultati}
\end{figure}

\begin{figure}[htpb]
\centering
\includegraphics[width=\textwidth]{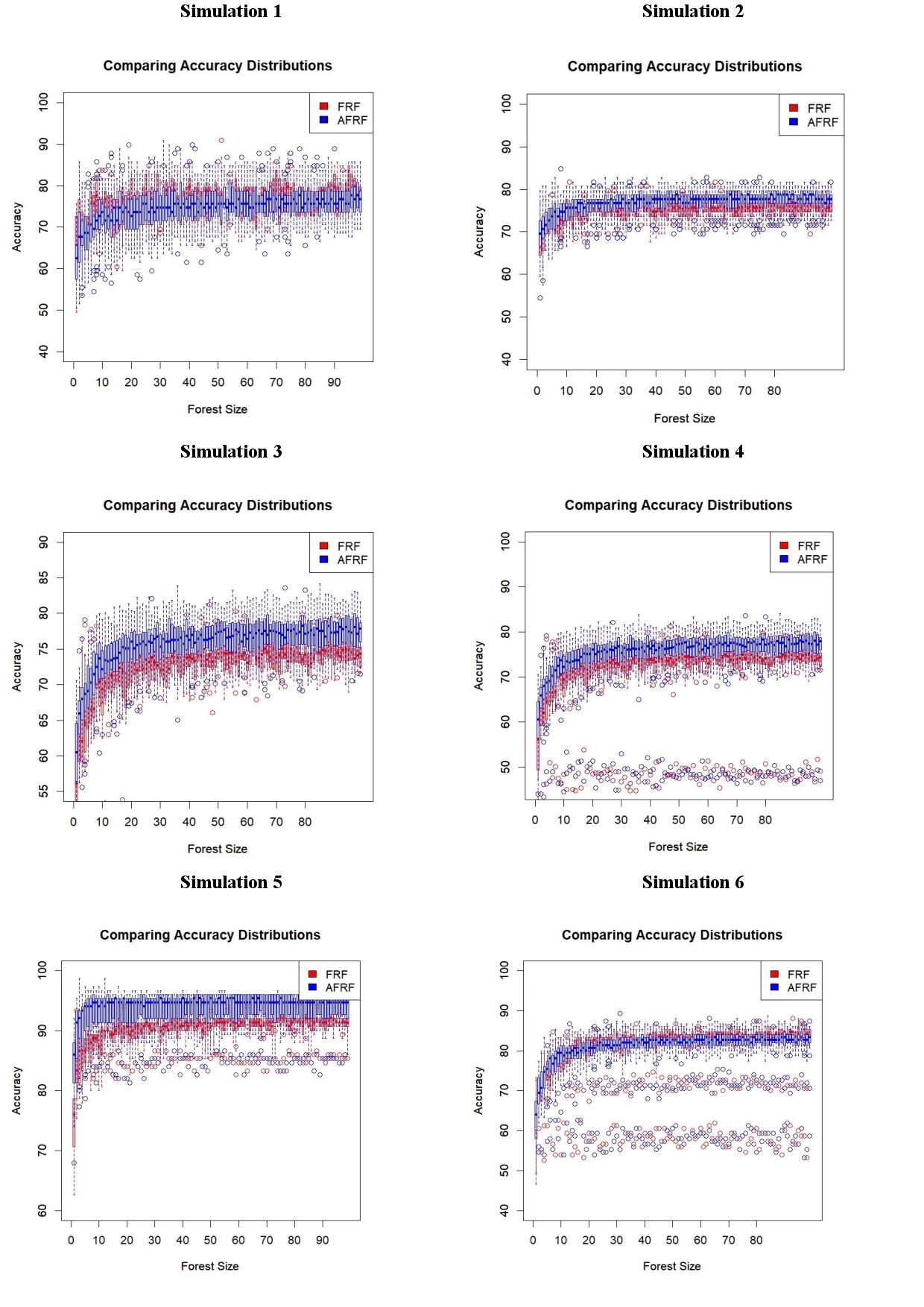}
\caption{Results of the simulated scenarios and comparison between FRF and AFRF in terms of distributions (to avoid claims of better results but due to chance).}
\label{Risultati2}
\end{figure}

\begin{figure}[htpb]
\centering
\includegraphics[width=\textwidth]{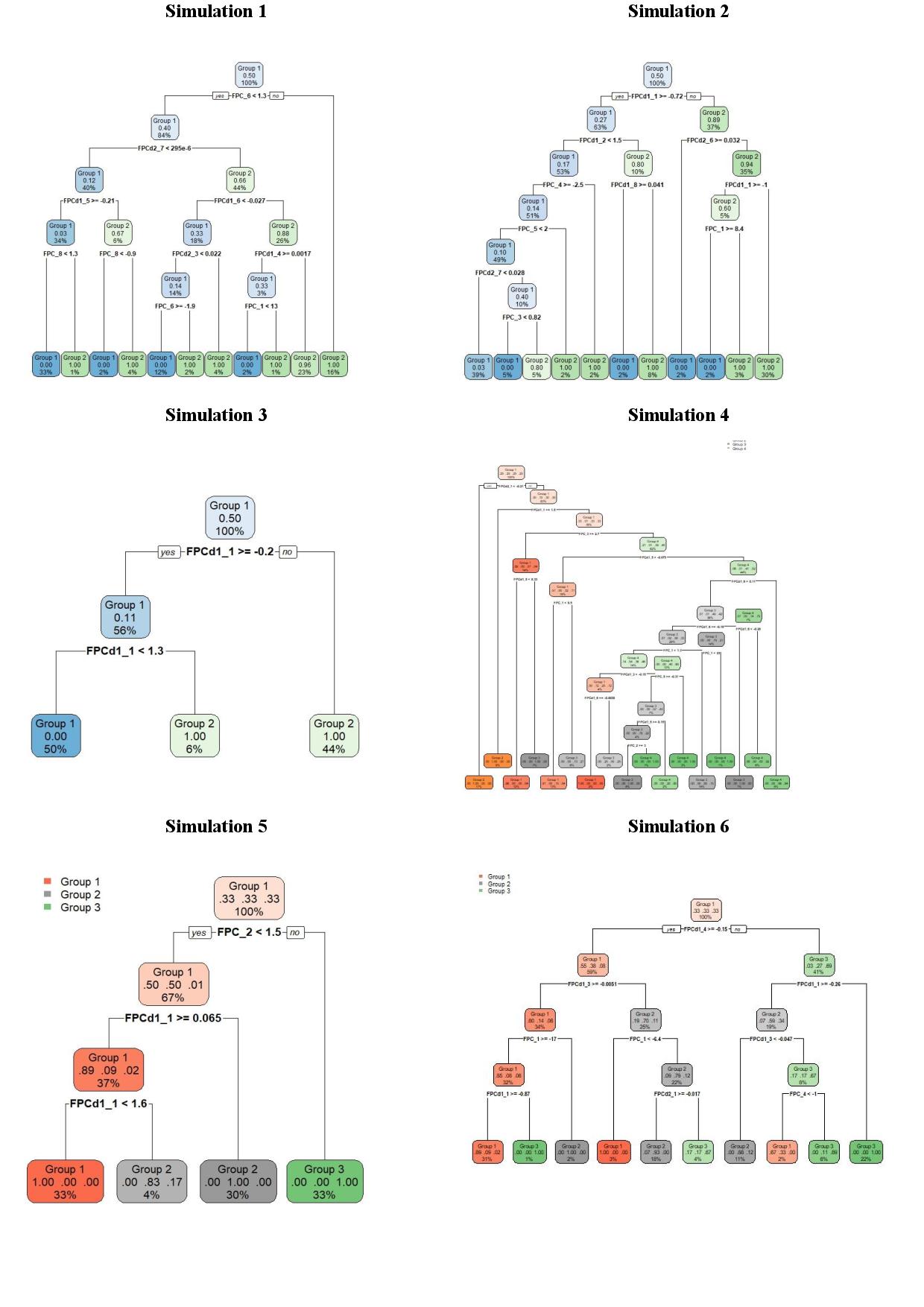}
\caption{Most representative AFCTs of the simulated scenarios.}
\label{Risultati3}
\end{figure}

Figure \ref{Risultati} depicts the outcomes of the six simulated scenarios and provides a comparative analysis between FRF and AFRF in terms of accuracy. Similar to the application on ECG200 data, we present the average accuracy across different numbers of augmented features for each forest size, offering a more reliable metric than relying on maximum accuracy, which might be susceptible to chance-induced fluctuations.
The results show that the proposed functional classifier (AFRF) consistently outperforms FRF in five simulated scenarios. This underscores the AFRF's robustness and efficacy across various scenarios.
In Figure \ref{Risultati2}, we present the accuracy distribution for each simulated scenario, offering insights into result variability and facilitating comparisons 
between FRF and AFRF regarding accuracy distributions across varying forest sizes. Each subplot corresponds to a different simulation, with the x-axis representing the forest size and the y-axis representing the accuracy. The boxplots illustrate the accuracy distribution for both FRF and AFRF. The comparison shows how the distributions evolve as the forest size increases, providing insight into the augmentation approach's performance stability and potential improvements. It is important to note that these comparisons are crucial to avoid attributing better results to chance rather than to the effectiveness of the AFRF methodology.
Finally, Figure \ref{Risultati3} illustrates the most representative AFCT (Augmented Functional Classification Tree) and elucidates how augmented features contribute to the classification rule. For a more in-depth understanding, additional details on the original functions, first and second derivatives for each simulated scenario can be found in the supplementary material.

\section{Discussion and Conclusions}
\label{sec5:discussion}

The field of supervised classification for curves is currently evolving. While there have been advancements, the integrated use of Functional Data Analysis (FDA) and tree-based classifiers still needs to be explored, with limited investigations conducted. Prior studies have investigated this combination from diverse methodological and applicative perspectives \citep{Yu_1999, Nerini_2007, Gregorutti_2015, Moller2016, ElHaouij2019, maturo2023supervised, Maturo2022SIM}. Despite these efforts, several aspects warrant further exploration and development.
Critical areas for improvement include enhancing the accuracy of functional classifiers by incorporating relevant features, introducing graphical tools to facilitate the interpretation of classification rules, conducting comprehensive simulation studies, and developing additional tools for determining optimal parameters in the supervised classification of functional data. The dynamic and promising nature of ongoing research in this domain presents numerous opportunities for further advancements and contributions to the evolving landscape of curve classification methodologies.

This paper presents an innovative supervised classification strategy combining Functional Data Analysis (FDA), tree-based ensembles, and extracting additional insights from curve analysis through diverse perspectives. The proposed approaches, namely Augmented Functional Classification Trees (AFCTs) and Augmented Functional Random Forests (AFRFs), are designed to tackle the challenges of learning from high-dimensional data, aiming to enhance classification performance.
By leveraging additional features derived from the original functions, such as sequential derivatives, the augmented functional data strategy captures nuanced information about the rate of change and local behaviour within the functional data. This augmentation process enriches the information provided to the functional classification tree ensembles, thereby enhancing the classifier's predictive power.
Additionally, we developed a novel feature importance metric designed explicitly for Augmented Functional Random Forests (AFRF) to address the limitations of conventional methods, such as Gini Importance and Permutation Importance when applied to functional data with derivatives. The inherent correlations among functional principal component scores derived from successive derivatives can bias traditional importance measures. To mitigate this, we introduced the Conditional Permutation Importance for Augmented Functional Principal Components (CPIAFPCs), which accurately assesses each score's contribution by conditioning on correlated features, providing a more reliable and interpretable model performance evaluation.

Extensive experimental evaluations on real-world and simulated datasets demonstrate the effectiveness of the proposed procedure. Comparisons with existing methods show exciting results in terms of classification performance. Hence, this research contributes to the field of FDA by demonstrating its potential for handling high-dimensional data in supervised classification tasks and providing valuable insights into the underlying functional characteristics of the data.
The results highlight the effectiveness of the AFCTs and AFRFs in improving classification performance and gaining a deeper understanding of the functional nature of the data. 
As demonstrated by the simulation studies, the results consistently affirm that adding additional features is a potent predictive tool for curves. Notably, the first simulated scenario exhibits no improvement over an approach based solely on the original curves. The rationale for this observation is straightforward, as depicted in Figure \ref{simulazioni}: the first scenario represents the simplest case, where the inclusion of additional features is unnecessary. In this context, both groups of curves exhibit straightforward trends and minimal variability, rendering the use of derivatives unnecessary.
Conversely, when considering scenarios with a greater number of classes and diverse curve shapes, the proposed classifier demonstrates substantial advantages. It adeptly captures local characteristics that the original curves struggle to discern with the same level of efficacy. This underscores the value of incorporating additional features, particularly in situations where the complexity of curve patterns demands a more nuanced approach for accurate classification.

This paper focused on FPCs decomposition as a powerful dimensional reduction tool for curves. Of course, the procedure can be extended to other functional classifiers and basis functions on a fixed-basis and data-driven basis. 
Future research directions in this area include further advancements in functional data analysis, exploring additional features for augmentation, and investigating models' explainability techniques for functional classifiers. 
The proposed methodology opens up new possibilities for functional data analysis and offers valuable insights into the underlying functional characteristics of the data. With continued research and exploration, we expect to uncover more innovative techniques for managing large quantities of functional data and interpreting the results from a statistical, causal, and inferential perspective.

\subsection*{Supplementary Information}

Supplementary files accompany this article.

\subsection*{Declarations}

The authors declare that they received no funding for this study and have no affiliations or involvement with any organization that has a financial or non-financial interest in the subject matter of this manuscript.

\subsection*{Funding and/or Conflicts of Interest/Competing Interests}

The authors confirm that they received no support from any organization for the submitted work. They also declare no affiliations or involvement with any organization or entity that has a financial or non-financial interest in the subject matter of this manuscript.

\subsection*{Use of Generative AI in Scientific Writing}

The authors used \textit{Grammarly AI} to improve the English language in preparing this work. They reviewed and edited the content as necessary and took full responsibility for the content of the publication.

 \subsection*{Data availability statement}
 
The authors utilized publicly available data for real-world applications. Simulation data can be provided upon request.

\bibliography{_biblio}


\begin{thebibliography}{31}
\ifx \bisbn   \undefined \def \bisbn  #1{ISBN #1}\fi
\ifx \binits  \undefined \def \binits#1{#1}\fi
\ifx \bauthor  \undefined \def \bauthor#1{#1}\fi
\ifx \batitle  \undefined \def \batitle#1{#1}\fi
\ifx \bjtitle  \undefined \def \bjtitle#1{#1}\fi
\ifx \bvolume  \undefined \def \bvolume#1{\textbf{#1}}\fi
\ifx \byear  \undefined \def \byear#1{#1}\fi
\ifx \bissue  \undefined \def \bissue#1{#1}\fi
\ifx \bfpage  \undefined \def \bfpage#1{#1}\fi
\ifx \blpage  \undefined \def \blpage #1{#1}\fi
\ifx \burl  \undefined \def \burl#1{\textsf{#1}}\fi
\ifx \doiurl  \undefined \def \doiurl#1{\url{https://doi.org/#1}}\fi
\ifx \betal  \undefined \def \betal{\textit{et al.}}\fi
\ifx \binstitute  \undefined \def \binstitute#1{#1}\fi
\ifx \binstitutionaled  \undefined \def \binstitutionaled#1{#1}\fi
\ifx \bctitle  \undefined \def \bctitle#1{#1}\fi
\ifx \beditor  \undefined \def \beditor#1{#1}\fi
\ifx \bpublisher  \undefined \def \bpublisher#1{#1}\fi
\ifx \bbtitle  \undefined \def \bbtitle#1{#1}\fi
\ifx \bedition  \undefined \def \bedition#1{#1}\fi
\ifx \bseriesno  \undefined \def \bseriesno#1{#1}\fi
\ifx \blocation  \undefined \def \blocation#1{#1}\fi
\ifx \bsertitle  \undefined \def \bsertitle#1{#1}\fi
\ifx \bsnm \undefined \def \bsnm#1{#1}\fi
\ifx \bsuffix \undefined \def \bsuffix#1{#1}\fi
\ifx \bparticle \undefined \def \bparticle#1{#1}\fi
\ifx \barticle \undefined \def \barticle#1{#1}\fi
\bibcommenthead
\ifx \bconfdate \undefined \def \bconfdate #1{#1}\fi
\ifx \botherref \undefined \def \botherref #1{#1}\fi
\ifx \url \undefined \def \url#1{\textsf{#1}}\fi
\ifx \bchapter \undefined \def \bchapter#1{#1}\fi
\ifx \bbook \undefined \def \bbook#1{#1}\fi
\ifx \bcomment \undefined \def \bcomment#1{#1}\fi
\ifx \oauthor \undefined \def \oauthor#1{#1}\fi
\ifx \citeauthoryear \undefined \def \citeauthoryear#1{#1}\fi
\ifx \endbibitem  \undefined \def \endbibitem {}\fi
\ifx \bconflocation  \undefined \def \bconflocation#1{#1}\fi
\ifx \arxivurl  \undefined \def \arxivurl#1{\textsf{#1}}\fi
\csname PreBibitemsHook\endcsname

\bibitem[\protect\citeauthoryear{Ramsay and Silverman}{2002}]{Ramsay2002}
\begin{bbook}
\bauthor{\bsnm{Ramsay}, \binits{J.O.}},
\bauthor{\bsnm{Silverman}, \binits{B.W.}}:
\bbtitle{Applied Functional Data Analysis: Methods and Case Studies}.
\bpublisher{Springer},
\blocation{New York, NY}
(\byear{2002}).
\doiurl{10.1007/b98886}
\end{bbook}
\endbibitem

\bibitem[\protect\citeauthoryear{Ferraty and Vieu}{2003}]{Ferraty2003}
\begin{barticle}
\bauthor{\bsnm{Ferraty}, \binits{F.}},
\bauthor{\bsnm{Vieu}, \binits{P.}}:
\batitle{Curves discrimination: a nonparametric functional approach}.
\bjtitle{Computational Statistics {\&} Data Analysis}
\bvolume{44}(\bissue{1-2}),
\bfpage{161}--\blpage{173}
(\byear{2003})
\doiurl{10.1016/s0167-9473(03)00032-x}
\end{barticle}
\endbibitem

\bibitem[\protect\citeauthoryear{Ramsay and Silverman}{2005}]{Ramsay2005}
\begin{bbook}
\bauthor{\bsnm{Ramsay}, \binits{J.}},
\bauthor{\bsnm{Silverman}, \binits{B.}}:
\bbtitle{Functional Data Analysis, 2nd Edn}.
\bpublisher{Springer},
\blocation{New York}
(\byear{2005}).
\doiurl{10.1007/b98888}
\end{bbook}
\endbibitem

\bibitem[\protect\citeauthoryear{Febrero-Bande and de~la Fuente}{2012}]{Febrero2012}
\begin{barticle}
\bauthor{\bsnm{Febrero-Bande}, \binits{M.}},
\bauthor{\bsnm{Fuente}, \binits{M.O.}}:
\batitle{Statistical computing in functional data analysis: The r package fda.usc}.
\bjtitle{Journal of Statistical Software}
\bvolume{51},
\bfpage{1}--\blpage{28}
(\byear{2012})
\doiurl{10.18637/jss.v051.i04}
\end{barticle}
\endbibitem

\bibitem[\protect\citeauthoryear{Ferraty and Vieu}{2006}]{Ferraty2006}
\begin{bbook}
\bauthor{\bsnm{Ferraty}, \binits{F.}},
\bauthor{\bsnm{Vieu}, \binits{P.}}:
\bbtitle{Nonparametric Functional Data Analysis}.
\bpublisher{Springer},
\blocation{New York}
(\byear{2006}).
\doiurl{10.1007/0-387-36620-2}
\end{bbook}
\endbibitem

\bibitem[\protect\citeauthoryear{Aguilera and Aguilera-Morillo}{2013}]{Aguilera2013}
\begin{barticle}
\bauthor{\bsnm{Aguilera}, \binits{A.}},
\bauthor{\bsnm{Aguilera-Morillo}, \binits{M.}}:
\batitle{Penalized {PCA} approaches for {B}-spline expansions of smooth functional data}.
\bjtitle{Applied Mathematics and Computation}
\bvolume{219},
\bfpage{7805}--\blpage{7819}
(\byear{2013})
\doiurl{10.1016/j.amc.2013.02.009}
\end{barticle}
\endbibitem

\bibitem[\protect\citeauthoryear{Maturo et~al.}{2020}]{maturo2019fuzzy}
\begin{barticle}
\bauthor{\bsnm{Maturo}, \binits{F.}},
\bauthor{\bsnm{Ferguson}, \binits{J.}},
\bauthor{\bsnm{{Di Battista}}, \binits{T.}},
\bauthor{\bsnm{Ventre}, \binits{V.}}:
\batitle{A fuzzy functional k-means approach for monitoring {I}talian regions according to health evolution over time}.
\bjtitle{Soft Computing}
\bvolume{24},
\bfpage{13741}--\blpage{13755}
(\byear{2020})
\doiurl{10.1007/s00500-019-04505-2}
\end{barticle}
\endbibitem

\bibitem[\protect\citeauthoryear{Cuevas}{2014}]{Cuevas2014}
\begin{barticle}
\bauthor{\bsnm{Cuevas}, \binits{A.}}:
\batitle{A partial overview of the theory of statistics with functional data}.
\bjtitle{Journal of Statistical Planning and Inference}
\bvolume{147},
\bfpage{1}--\blpage{23}
(\byear{2014})
\doiurl{10.1016/j.jspi.2013.04.002}
\end{barticle}
\endbibitem

\bibitem[\protect\citeauthoryear{Cuevas et~al.}{2007}]{Cuevas_2007}
\begin{barticle}
\bauthor{\bsnm{Cuevas}, \binits{A.}},
\bauthor{\bsnm{Febrero}, \binits{M.}},
\bauthor{\bsnm{Fraiman}, \binits{R.}}:
\batitle{Robust estimation and classification for functional data via projection-based depth notions}.
\bjtitle{Computational Statistics}
\bvolume{22}(\bissue{3}),
\bfpage{481}--\blpage{496}
(\byear{2007})
\doiurl{10.1007/s00180-007-0053-0}
\end{barticle}
\endbibitem

\bibitem[\protect\citeauthoryear{Preda et~al.}{2007}]{Preda_2007}
\begin{barticle}
\bauthor{\bsnm{Preda}, \binits{C.}},
\bauthor{\bsnm{Saporta}, \binits{G.}},
\bauthor{\bsnm{L{\'{e}}v{\'{e}}der}, \binits{C.}}:
\batitle{{PLS} classification of functional data}.
\bjtitle{Computational Statistics}
\bvolume{22}(\bissue{2}),
\bfpage{223}--\blpage{235}
(\byear{2007})
\doiurl{10.1007/s00180-007-0041-4}
\end{barticle}
\endbibitem

\bibitem[\protect\citeauthoryear{Aguilera-Morillo et~al.}{2012}]{AguileraMorillo2012}
\begin{barticle}
\bauthor{\bsnm{Aguilera-Morillo}, \binits{M.}},
\bauthor{\bsnm{Aguilera}, \binits{A.}},
\bauthor{\bsnm{Escabias}, \binits{M.}},
\bauthor{\bsnm{Valderrama}, \binits{M.J.}}:
\batitle{Penalized spline approaches for functional logit regression}.
\bjtitle{Test}
\bvolume{22}(\bissue{2}),
\bfpage{251}--\blpage{277}
(\byear{2012})
\doiurl{10.1007/s11749-012-0307-1}
\end{barticle}
\endbibitem

\bibitem[\protect\citeauthoryear{Escabias et~al.}{2014}]{Escabias2014}
\begin{barticle}
\bauthor{\bsnm{Escabias}, \binits{M.}},
\bauthor{\bsnm{Aguilera}, \binits{A.M.}},
\bauthor{\bsnm{Aguilera-Morillo}, \binits{M.C.}}:
\batitle{Functional {PCA} and base-line logit models}.
\bjtitle{Journal of Classification}
\bvolume{31}(\bissue{3}),
\bfpage{296}--\blpage{324}
(\byear{2014})
\doiurl{10.1007/s00357-014-9162-y}
\end{barticle}
\endbibitem

\bibitem[\protect\citeauthoryear{Gregorutti et~al.}{2015}]{Gregorutti_2015}
\begin{barticle}
\bauthor{\bsnm{Gregorutti}, \binits{B.}},
\bauthor{\bsnm{Michel}, \binits{B.}},
\bauthor{\bsnm{Saint-Pierre}, \binits{P.}}:
\batitle{Grouped variable importance with random forests and application to multiple functional data analysis}.
\bjtitle{Computational Statistics {\&} Data Analysis}
\bvolume{90},
\bfpage{15}--\blpage{35}
(\byear{2015})
\doiurl{10.1016/j.csda.2015.04.002}
\end{barticle}
\endbibitem

\bibitem[\protect\citeauthoryear{Yu and Lambert}{1999}]{Yu_1999}
\begin{barticle}
\bauthor{\bsnm{Yu}, \binits{Y.}},
\bauthor{\bsnm{Lambert}, \binits{D.}}:
\batitle{Fitting trees to functional data, with an application to time-of-day patterns}.
\bjtitle{Journal of Computational and Graphical Statistics}
\bvolume{8}(\bissue{4}),
\bfpage{749}--\blpage{762}
(\byear{1999})
\doiurl{10.1080/10618600.1999.10474847}
\end{barticle}
\endbibitem

\bibitem[\protect\citeauthoryear{Nerini and Ghattas}{2007}]{Nerini_2007}
\begin{barticle}
\bauthor{\bsnm{Nerini}, \binits{D.}},
\bauthor{\bsnm{Ghattas}, \binits{B.}}:
\batitle{Classifying densities using functional regression trees: Applications in oceanology}.
\bjtitle{Computational Statistics {\&} Data Analysis}
\bvolume{51}(\bissue{10}),
\bfpage{4984}--\blpage{4993}
(\byear{2007})
\doiurl{10.1016/j.csda.2006.09.028}
\end{barticle}
\endbibitem

\bibitem[\protect\citeauthoryear{M{\"{o}}ller et~al.}{2016}]{Moller2016}
\begin{barticle}
\bauthor{\bsnm{M{\"{o}}ller}, \binits{A.}},
\bauthor{\bsnm{Tutz}, \binits{G.}},
\bauthor{\bsnm{Gertheiss}, \binits{J.}}:
\batitle{{Random forests for functional covariates}}.
\bjtitle{Journal of Chemometrics}
(\byear{2016})
\doiurl{10.1002/cem.2849}
\end{barticle}
\endbibitem

\bibitem[\protect\citeauthoryear{Haouij et~al.}{2018}]{El_Haouij_2018}
\begin{barticle}
\bauthor{\bsnm{Haouij}, \binits{N.E.}},
\bauthor{\bsnm{Poggi}, \binits{J.-M.}},
\bauthor{\bsnm{Ghozi}, \binits{R.}},
\bauthor{\bsnm{Sevestre-Ghalila}, \binits{S.}},
\bauthor{\bsnm{Jaïdane}, \binits{M.}}:
\batitle{Random forest-based approach for physiological functional variable selection for driver's stress level classification}.
\bjtitle{Statistical Methods {\&} Applications}
\bvolume{28}(\bissue{1}),
\bfpage{157}--\blpage{185}
(\byear{2018})
\doiurl{10.1007/s10260-018-0423-5}
\end{barticle}
\endbibitem

\bibitem[\protect\citeauthoryear{Rahman et~al.}{2019}]{Rahman_2019}
\begin{botherref}
\oauthor{\bsnm{Rahman}, \binits{R.}},
\oauthor{\bsnm{Dhruba}, \binits{S.}},
\oauthor{\bsnm{Ghosh}, \binits{S.}},
\oauthor{\bsnm{Pal}, \binits{R.}}:
Functional random forest with applications in dose-response predictions.
Scientific Reports
\textbf{9}(1)
(2019)
\doiurl{10.1038/s41598-018-38231-w}
\end{botherref}
\endbibitem

\bibitem[\protect\citeauthoryear{Maturo and Verde}{2023}]{maturo2023supervised}
\begin{barticle}
\bauthor{\bsnm{Maturo}, \binits{F.}},
\bauthor{\bsnm{Verde}, \binits{R.}}:
\batitle{Supervised classification of curves via a combined use of functional data analysis and tree-based methods}.
\bjtitle{Computational Statistics}
\bvolume{38}(\bissue{1}),
\bfpage{419}--\blpage{459}
(\byear{2023})
\doiurl{10.1007/s00180-022-01236-1}
\end{barticle}
\endbibitem

\bibitem[\protect\citeauthoryear{Maturo and Verde}{2024}]{maturo2024combining}
\begin{barticle}
\bauthor{\bsnm{Maturo}, \binits{F.}},
\bauthor{\bsnm{Verde}, \binits{R.}}:
\batitle{Combining unsupervised and supervised learning techniques for enhancing the performance of functional data classifiers}.
\bjtitle{Computational Statistics}
\bvolume{39}(\bissue{1}),
\bfpage{239}--\blpage{270}
(\byear{2024})
\doiurl{10.1007/s00180-022-01259-8}
\end{barticle}
\endbibitem

\bibitem[\protect\citeauthoryear{Maturo and Verde}{2022}]{Maturo2022SIM}
\begin{barticle}
\bauthor{\bsnm{Maturo}, \binits{F.}},
\bauthor{\bsnm{Verde}, \binits{R.}}:
\batitle{Pooling random forest and functional data analysis for biomedical signals supervised classification: theory and application to electrocardiogram data}.
\bjtitle{Statistics in Medicine}
\bvolume{41},
\bfpage{2247}--\blpage{2275}
(\byear{2022})
\doiurl{10.1002/sim.9353}
\end{barticle}
\endbibitem

\bibitem[\protect\citeauthoryear{Hastie et~al.}{2009}]{Hastie2009}
\begin{bbook}
\bauthor{\bsnm{Hastie}, \binits{T.}},
\bauthor{\bsnm{Tibshirani}, \binits{R.}},
\bauthor{\bsnm{Friedman}, \binits{J.}}:
\bbtitle{The Elements of Statistical Learning: Data Mining, Inference, and Prediction},
\bedition{2nd} edn.
\bsertitle{Springer Series in Statistics}.
\bpublisher{Springer},
\blocation{New York, NY}
(\byear{2009}).
\doiurl{10.1007/978-0-387-84858-7}
\end{bbook}
\endbibitem

\bibitem[\protect\citeauthoryear{Ramsay et~al.}{2009}]{Ramsay2009}
\begin{bchapter}
\bauthor{\bsnm{Ramsay}, \binits{J.O.}},
\bauthor{\bsnm{Hooker}, \binits{G.}},
\bauthor{\bsnm{Graves}, \binits{S.}}:
\bctitle{Introduction to functional data analysis}.
In: \bbtitle{Functional Data Analysis with R and {MATLAB}},
pp. \bfpage{1}--\blpage{19}.
\bpublisher{Springer}, \blocation{???}
(\byear{2009}).
\doiurl{10.1007/978-0-387-98185-7\_1}
\end{bchapter}
\endbibitem

\bibitem[\protect\citeauthoryear{Hyafil and Rivest}{1976}]{Hyafil1976}
\begin{barticle}
\bauthor{\bsnm{Hyafil}, \binits{L.}},
\bauthor{\bsnm{Rivest}, \binits{R.L.}}:
\batitle{{Constructing optimal binary decision trees is NP-complete}}.
\bjtitle{Information Processing Letters}
(\byear{1976})
\doiurl{10.1016/0020-0190(76)90095-8}
\end{barticle}
\endbibitem

\bibitem[\protect\citeauthoryear{Quinlan}{1986}]{Quinlan1986}
\begin{barticle}
\bauthor{\bsnm{Quinlan}, \binits{J.R.}}:
\batitle{{Induction of Decision Trees}}.
\bjtitle{Machine Learning}
(\byear{1986})
\doiurl{10.1023/A:1022643204877}
\end{barticle}
\endbibitem

\bibitem[\protect\citeauthoryear{Therneau and Atkinson}{2019}]{rpart}
\begin{botherref}
\oauthor{\bsnm{Therneau}, \binits{T.}},
\oauthor{\bsnm{Atkinson}, \binits{B.}}:
Rpart: Recursive Partitioning and Regression Trees.
(2019).
\doiurl{https://CRAN.R-project.org/package=rpart} .
R package version 4.1-15
\end{botherref}
\endbibitem

\bibitem[\protect\citeauthoryear{Breiman}{1996}]{Breiman1996}
\begin{barticle}
\bauthor{\bsnm{Breiman}, \binits{L.}}:
\batitle{Bagging predictors}.
\bjtitle{Machine Learning}
(\byear{1996})
\doiurl{10.1007/bf00058655}
\end{barticle}
\endbibitem

\bibitem[\protect\citeauthoryear{Ho}{1998}]{Ho1998}
\begin{barticle}
\bauthor{\bsnm{Ho}, \binits{T.K.}}:
\batitle{{The random subspace method for constructing decision forests}}.
\bjtitle{IEEE Transactions on Pattern Analysis and Machine Intelligence}
(\byear{1998})
\doiurl{10.1109/34.709601}
\end{barticle}
\endbibitem

\bibitem[\protect\citeauthoryear{Strobl et~al.}{2008}]{strobl2008conditional}
\begin{barticle}
\bauthor{\bsnm{Strobl}, \binits{C.}},
\bauthor{\bsnm{Boulesteix}, \binits{A.-L.}},
\bauthor{\bsnm{Kneib}, \binits{T.}},
\bauthor{\bsnm{Augustin}, \binits{T.}},
\bauthor{\bsnm{Zeileis}, \binits{A.}}:
\batitle{Conditional variable importance for random forests}.
\bjtitle{BMC Bioinformatics}
\bvolume{9}(\bissue{1}),
\bfpage{1}--\blpage{11}
(\byear{2008})
\doiurl{10.1186/1471-2105-9-307}
\end{barticle}
\endbibitem

\bibitem[\protect\citeauthoryear{Olszewski}{2001}]{ecg200dataset}
\begin{botherref}
\oauthor{\bsnm{Olszewski}, \binits{R.}}:
ECG200 Dataset.
Carnegie Mellon University.
Accessed: 2024-08-22
(2001).
\url{https://www.timeseriesclassification.com/description.php?Dataset=ECG200}
\end{botherref}
\endbibitem

\bibitem[\protect\citeauthoryear{{El Haouij} et~al.}{2019}]{ElHaouij2019}
\begin{barticle}
\bauthor{\bsnm{{El Haouij}}, \binits{N.}},
\bauthor{\bsnm{Poggi}, \binits{J.M.}},
\bauthor{\bsnm{Ghozi}, \binits{R.}},
\bauthor{\bsnm{Sevestre-Ghalila}, \binits{S.}},
\bauthor{\bsnm{Ja{\"{i}}dane}, \binits{M.}}:
\batitle{{Random forest-based approach for physiological functional variable selection for driver's stress level classification}}.
\bjtitle{Statistical Methods and Applications}
(\byear{2019})
\doiurl{10.1007/s10260-018-0423-5}
\end{barticle}
\endbibitem

\end{thebibliography}

\end{document}